\documentclass[journal,10pt,twoside]{IEEEtran}
\ifCLASSINFOpdf
\usepackage[pdftex]{graphicx}
\graphicspath{{../pdf/}{../jpeg/}}
\DeclareGraphicsExtensions{.pdf,.jpeg,.png}
\else 
\usepackage[dvips]{graphicx}
\graphicspath{{../eps/}}
\DeclareGraphicsExtensions{.eps}
\fi
\usepackage{epstopdf}
\usepackage{amsmath,amsthm, amssymb}
\usepackage{threeparttable}
\usepackage{bm}
\usepackage{multirow}
\usepackage{booktabs}

\usepackage{color}
\renewcommand{\textcolor}[2]{#2}
\usepackage{tabularx}
\usepackage{setspace}
\usepackage{verbatim}
\usepackage{stfloats}
\usepackage{caption}
\usepackage{float}
\usepackage[T1]{fontenc}
\usepackage{soul}
\usepackage{url}
\usepackage[utf8]{inputenc}
\usepackage{cite}
\usepackage[hidelinks]{hyperref} 
\usepackage{stfloats}
\usepackage{mathtools}
\usepackage{amsmath}
\usepackage{amsthm}
\usepackage{amssymb}
\usepackage{amsfonts}
\allowdisplaybreaks[0]
\usepackage{subfig}
\usepackage{fancyhdr}
\usepackage{cancel}
\usepackage{units}

\makeatletter
\newif\if@restonecol
\makeatother

\usepackage[linesnumbered,ruled,vlined]{algorithm2e}
	\usepackage{algpseudocode}
	\usepackage{amsmath}

\usepackage[switch]{lineno}

\begin{document}
		\title{Graph Neural Networks and Deep Reinforcement Learning Based Resource Allocation for V2X Communications}
		
		\author{
			Maoxin Ji, Qiong Wu, ~\IEEEmembership{Senior Member,~IEEE}, Pingyi Fan, ~\IEEEmembership{Senior Member,~IEEE},\\ Nan Cheng, ~\IEEEmembership{Senior Member,~IEEE}, Wen Chen, ~\IEEEmembership{Senior Member,~IEEE},\\ Jiangzhou Wang, ~\IEEEmembership{Fellow,~IEEE}, Khaled B. Letaief, ~\IEEEmembership{Fellow,~IEEE}
			\thanks{This work was supported in part by the National Natural Science Foundation of China under Grant 61701197 and Grant 62071296; in part by the National Key Research and Development Program of China under Grant 2021YFA1000500(4); in part by the National Key Project under Grant 2020YFB1807700; in part by Shanghai Kewei under Grant 22JC1404000; in part by the Research Grants Council under the Areas of Excellence Scheme under Grant AoE/E-601/22-R; and in part by the 111 Project under Grant B23008.}
			\thanks{Maoxin Ji and Qiong Wu are with the School of Internet of Things Engineering, Jiangnan University, Wuxi 214122, China (e-mail: maoxinji@stu.jiangnan.edu.cn, qiongwu@jiangnan.edu.cn).}
			\thanks{Pingyi Fan is with the Department of Electronic Engineering, Beijing National Research Center for Information Science and Technology, Tsinghua University, Beijing 100084, China (e-mail: fpy@tsinghua.edu.cn).}
			\thanks{Nan Cheng is with the State Key Lab. of ISN and School of Telecommunications Engineering, Xidian University, Xi'an 710071, China (e-mail: dr.nan.cheng@ieee.org).}
			\thanks{Wen Chen is with the Department of Electronic Engineering, Shanghai Jiao Tong University, Shanghai 200240, China (e-mail: wenchen@sjtu.edu.cn).}
			\thanks{Jiangzhou Wang is with the School of Engineering, University of Kent, CT2 7NT Canterbury, U.K. (email: j.z.wang@kent.ac.uk).}
			\thanks{K. B. Letaief is with the Department of Electrical and Computer Engineering, the Hong Kong University of Science and Technology (HKUST), Hong Kong (email: eekhaled@ust.hk).}
			\thanks{Copyright (c) 2024 IEEE. Personal use of this material is permitted. However, permission to use this material for any other purposes must be obtained from the IEEE by sending a request to pubs-permissions@ieee.org.}
		}
		
		
		
		\maketitle

	\begin{abstract}
		In the rapidly evolving landscape of Internet of Vehicles (IoV) technology, Cellular Vehicle-to-Everything (C-V2X) communication has attracted much attention due to its superior performance in coverage, latency, and throughput. Resource allocation within C-V2X is crucial for ensuring the transmission of safety information and meeting the stringent requirements for ultra-low latency and high reliability in Vehicle-to-Vehicle (V2V) communication. This paper proposes a method that integrates Graph Neural Networks (GNN) with Deep Reinforcement Learning (DRL) to address this challenge. By constructing a dynamic graph with communication links as nodes and employing the Graph Sample and Aggregation (GraphSAGE) model to adapt to changes in graph structure, the model aims to ensure a high success rate for V2V communication while minimizing interference on Vehicle-to-Infrastructure (V2I) links, thereby ensuring the successful transmission of V2V link information and maintaining high transmission rates for V2I links. The proposed method retains the global feature learning capabilities of GNN and supports distributed network deployment, allowing vehicles to extract low-dimensional features that include structural information from the graph network based on local observations and to make independent resource allocation decisions. Simulation results indicate that the introduction of GNN, with a modest increase in computational load, effectively enhances the decision-making quality of agents, demonstrating superiority to other methods. This study not only provides a theoretically efficient resource allocation strategy for V2V and V2I communications but also paves a new technical path for resource management in practical IoV environments.
	\end{abstract}


	\IEEEpeerreviewmaketitle
	
	\section{Introduction}
	\label{sec1}
	\IEEEPARstart{I}{n} the development of smart cities, Vehicle-to-Everything (V2X) technology plays an essential role within intelligent transportation systems \cite{6740844, Feng2024JointAB, Luan2023RobustBD}. V2X is designed to enable comprehensive communication between vehicles and their surrounding environments, including other vehicles, transportation infrastructure, pedestrians, and network resources \cite{7513432, 5888501, Wu2024Characterizing}. As the automotive industry transitions toward cutting-edge technologies such as autonomous driving, intelligent navigation, and automated parking \cite{6823640, 8058008, 8584062, Luo2024EdgeCooper, Wang2024ValueMatters}, the importance of V2X is increasingly underscored. However, V2X still faces challenges in terms of performance and safety.
	
	\textcolor{red}{Given these challenges, various V2X technologies have been developed to address different aspects of vehicular communication \cite{10636300, 10453965}.} Among the various V2X technologies, Cellular Vehicle-to-Everything (C-V2X) is considered to offer higher data rates, lower latency, and higher reliability than IEEE 802.11p technology. Recently, the 3rd Generation Partnership Project (3GPP) has standardized the New Radio Vehicle-to-Everything (NR-V2X) technology in its Release 16 standards \cite{9345798, 9734746}.
	
	\textcolor{red}{However, with the deployment of these advanced technologies, new issues arise \cite{10643168, 10536013, 5586664, 10163760, 9259112}.} In practical deployments, the resource allocation problem becomes a crucial aspect of vehicular network technology to support the substantial wireless communication demands brought by V2X communication \cite{9088326, Zhang2024Generative, Wu2023Characterizing, Zhuang2020SDN_NFV, Chen2023JointOS, 8672604, 9575181}. This problem is typically NP-hard, making it challenging to simultaneously meet the reliability requirements of vehicle-to-vehicle (V2V) links and the rate requirements of vehicle-to-infrastructure (V2I) links in vehicular network environments \cite{8318569, Zhang2023Energy}. Additionally, traditional resource allocation methods often rely on accurate channel state information (CSI) \cite{10620366}, which is quite difficult in environments where vehicles move at high speeds.
	
	\textcolor{red}{To tackle these resource allocation challenges, researchers have turned to advanced algorithms.} With the advancement of deep learning (DL) and reinforcement learning (RL), researchers have begun to leverage the powerful function approximation characteristics of deep reinforcement learning (DRL) to address the resource allocation problem \cite{9507541, 8730522, Lin2020Dynamic, Lin2021Contour, 10387423, 10654286}. In the presence of inaccurate CSI, DRL can accumulate experience through continuous trial and error, learning a general strategy that can achieve better performance in distributed resource allocation scenarios.

	Nonetheless, it has been observed that, under normal circumstances, centralized resource allocation schemes typically outperform their distributed counterparts in terms of effectiveness \cite{Shen2024RingSFL}. This superiority is attributed to the centralized approach's access to global information on resource occupancy and the specific demands of each link, which facilitates the formulation of superior allocation strategies.
	
	In a distributed setting, however, the agents within reinforcement learning are limited to making decisions based on their local observations, which often lack sufficient total system information. This limitation hinders effective collaboration among vehicles, significantly compromising the quality of global resource scheduling. The scenario is further complicated by the high likelihood of resource collisions and congestion, as well as the difficulty in managing mutual interference between communication links. Additionally, in environments where vehicles are in high-speed motion, the CSI derived from local vehicle observations is laden with considerable noise \cite{8638940}, making it challenging to accurately characterize the channel conditions. This, in turn, can adversely affect the performance of DRL.
	
	\textcolor{red}{To address these challenges, researchers have explored the potential of Graph Neural Networks (GNNs).} GNNs possess unique advantages in extracting global information and mitigating noise, which is reflected in their superior performance across a variety of tasks such as link prediction \cite{Zhang2018LinkPB}, node classification \cite{Kipf2016SemiSupervisedCW}, graph classification \cite{Zhang2018AnED}, and graph generation. The mechanism of GNNs involves iteratively aggregating features from neighboring nodes, allowing information to propagate layer by layer along the edges. Consequently, multi-layer GNNs are capable of capturing node features that encapsulate global information while also effectively suppressing noise during the aggregation process. This hierarchical and iterative approach to information processing enables GNNs to discern complex patterns and dependencies within graph-structured data, thereby enhancing their utility in a wide range of applications.
	
	In fact, in the literature, many works have already attempted to integrate GNNs with DRL to address complex issues. However, due to the unique structure of GNNs, most methods that combine GNNs operate as a centralized framework. Moreover, traditional GNNs struggle to handle dynamic changes in graph structures; once the number of nodes or the connections between nodes change, various adjustments to the graph are required, and in some cases, the network needs to be retrained. This is particularly problematic in vehicular networks where vehicles are in high-speed motion and the number of vehicles frequently fluctuates, as it can be detrimental to the network's adaptability and efficiency.
	
	The Graph Sample and Aggregate (GraphSAGE) model, introduced by William Hamilton, Rex Ying, and Jure Leskovec in 2017 \cite{Hamilton2017InductiveRL}, is a variant of GNNs tailored for dynamic graph structures. \textcolor{red}{It can learn a general aggregation function applicable to all nodes to handle the issue of changing node quantities, while also avoiding high computational costs from a large number of neighbors by sampling nodes at each layer.} However, GraphSAGE primarily focuses on aggregating node features and overlooks the significance of edge weights. In the context of vehicular networks, where V2V links are represented as nodes and the mutual interference between these links is depicted as edges, \textcolor{red}{the weight of edges can intuitively represent the strength of the interference, thus becoming a critical parameter}.

	Moreover, the vehicular network environment is commonly represented as a complete graph, where nodes(V2V Links) are interconnected through the relationships of interference. As the number of vehicles in the environment grows, the graph's complexity escalates, leading to a substantial increase in the demand for computational resources. This high resource consumption is not feasible in a vehicular network environment where low latency is a critical requirement.
	
	\textcolor{red}{In this paper, to achieve a more efficient distributed resource allocation and address the influence of limited and inaccurate local observation states in vehicles}, we propose a GNN-assisted distributed deep reinforcement learning (DRL) algorithm\footnote{\href{https://github.com/qiongwu86/GNN-and-DRL-Based-Resource-Allocation-for-V2X-Communications}{The source code has been released at: https://github.com/qiongwu86/GNN-and-DRL-Based-Resource-Allocation-for-V2X-Communications}}. The main contributions of this paper are summarized as follows:
	
	\begin{itemize}
		\item[1)] To enrich the state information available for vehicles to autonomously select resources, we introduce the GraphSAGE framework, which is capable of adapting to dynamic graph structures, and we take into account the characteristics of edges.
		
		\item[2)] To construct the \textcolor{red}{GraphSAGE} framework, and find a way to do the resource allocation, we integrate GNN with Double Deep Q-Network (DDQN) in a distributed scenario, where each V2V link is represented as a node in the graph and as an agent in the DDQN. In addition, the computational complexity and the robustness to the increasing of vehicles. We then test this approach by simulation.
	\end{itemize}
	
	The remainder of this paper is organized as follows: Section \ref{sec2} reviews the related work. Section \ref{sec3} introduces the system model. Section \ref{sec4} presents the proposed GNN model, providing a novel graph construction method. In Section \ref{sec5}, we utilize the previously constructed GNN model to extract low-dimensional features containing global information based on local observations of vehicles and collaborate with the DDQN model to solve the joint resource allocation problem. Section \ref{sec6} presents the simulation results. Finally, Section \ref{sec7} concludes the paper.\textcolor{red}{To better understand the proposed method, we will next review the latest developments in the relevant directions.}
	
	\section{Related Work}
	\label{sec2}
 	In this section, we first review some related work based on traditional methods, and then we review the latest work that utilizes DRL and GNN to address the resource allocation problem.
	\subsection{\textcolor{red}{Traditional Methods}}
	Traditional resource allocation methods often employ game theory, auction theory, and evolutionary algorithms for their studies. In \cite{7270335}, a wireless channel resource management method based on Device-to-Device (D2D) communication was investigated. \textcolor{red}{The requirements for delay and reliability in V2V communication} were transformed into optimization constraints. \textcolor{red}{These constraints can be directly computed using slowly varying channel information.} The channel resource allocation problem was formulated as an optimization problem. A heuristic algorithm was employed to solve it. 
	In \cite{8331869}, vehicle service access categories were divided into safety and non-safety Vehicle User Equipment. Its target was to maximize the system's total throughput.
	In \cite{8374081}, coexistence issues between C-V2X users and vehicle ad hoc network users in unlicensed spectrum were studied. \textcolor{red}{A spectrum sharing scheme based on energy sensing was proposed, in order to reduce conflicts between the two types of users, where matching theory was used to study the wireless resource allocation problem, and the dynamic resource allocation algorithm for vehicles was presented.} 
	In \cite{8993812}, the exact expression for the ergodic capacity of a single VUE was derived, considering imperfect CSI. A Simulated Annealing (SA) algorithm was used to obtain a good power allocation result. 
	In \cite{8490729}, a low-latency V2V communication resource allocation scheme was proposed based on 802.11p and C-V2X technologies. \textcolor{red}{Cellular base stations were used to determine channel selection in a hybrid architecture. The delay minimization problem was formulated as a maximum weighted independent set problem. A greedy V2V link selection algorithm was proposed, with a derived theoretical performance lower bound.}
	
	In \cite{4600213}, a multi-state cooperation methodology was proposed. Energy was allocated among nodes state-by-state using geometric and network decomposition methods. \textcolor{red}{The duration of each state was optimized to maximize the utility of the nodes.}
	In \cite{1313070}, a cross-layer adaptive resource allocation algorithm for packet-based \textcolor{red}{Orthogonal Frequency Division Multiplexing (OFDM)} systems was presented. This algorithm considered random traffic arrivals and user fairness issues. It enhanced system spectral efficiency and improved queuing performance.
	In \cite{7889030}, a wireless powered communication network with cooperative groups was investigated. A semi-definite relaxation method was employed to jointly optimize time allocation, beamforming vectors, and power distribution. \textcolor{red}{This was done under the constraints of available power and quality of service requirements for two communication groups.} The global optimal solution for each problem was guaranteed.
	In \cite{7488201}, a scenario in sensor networks was studied where a mobile control center powered the sensors via RF signals and collected their information. Two solutions were provided, which enhanced energy efficiency. 
	\textcolor{red}{In \cite{4489632}, the multicast routing problem based on network coding was explored. The aim was to achieve the maximum flow multicast routes in ad hoc networks.} \textcolor{red}{Statistical characteristics of encoding nodes and maximum flow in ad hoc networks were analyzed based on random graph theory.}
	In \cite{1207121}, node mobility was considered. A method capable of tracking dynamic topology changes was proposed. \textcolor{red}{This method allowed for intersecting paths and significantly improved end-to-end delay and packet delivery ratio performance.}
	\subsection{\textcolor{red}{Methods Based on DRL and GNN}}
 	Recently, DRL and GNN have increasingly been applied to solve resource allocation problems. In \cite{8633948}, reinforcement learning was first introduced to the resource allocation domain, proposing a distributed decision-making scheme. In \cite{9044821}, DRL was used to study resource allocation in an uplink Non-Orthogonal Multiple Access (NOMA) system, designing a discretized multi-Deep Q-learning (DQN) structure to reduce the output dimension, while introducing Deep Deterministic Policy Gradient (DDPG) for decision-making in continuous power allocation. In \cite{8944302}, considering user privacy protection, Federated Learning (FL) was introduced into resource allocation. On a large time scale, vehicles with similar channel conditions are grouped using spectral clustering, and vehicles within each group train models through FL. On a smaller time scale, vehicles upload their models to a central entity, which aggregates the models and distributes them back to each vehicle. In \cite{Yuan2021MetaReinforcementLB}, DQN and DDPG methods were applied within an Actor-Critic framework, using DQN for discrete channel selection and DDPG for continuous power selection. Finally, Meta Reinforcement Learning was used to learn a good initialization method for the network, enhancing its rapid adaptability in dynamically changing environments. In \cite{9285223}, graph embedding methods were used for D2D network link scheduling, using graph representation learning to extract node features for power selection, and analyzing the impact of supervised and unsupervised training on network performance. In \cite{9462385}, D2D communication links were treated as nodes in a graph, with interference between links represented as edges, modeling the wireless network as a directed graph for joint channel and power allocation, achieving near-optimal results with only a few samples, and requiring execution times of just a few milliseconds. In \cite{9505260}, a heterogeneous graph was constructed, and a heterogeneous GNN was used to learn policies, providing prior knowledge for Deep Neural Network(DNN) to make channel and power selections, reducing training complexity when the environment changes and DNN needs to be retrained. A parameter sharing strategy was also proposed to ensure that the prior knowledge learned by GNN is beneficial for DNN decision-making.
	
	 In summary, while much of the existing work has focused on refining network architectures to achieve more efficient resource allocation strategies, there has been a notable absence of consideration for enriching the information derived from local observations of vehicles. This gap in the literature has prompted us to embark on this research endeavor.
	\section{System Model}
	\label{sec3}
	\begin{figure}
	\centering
	\includegraphics[width=\columnwidth]{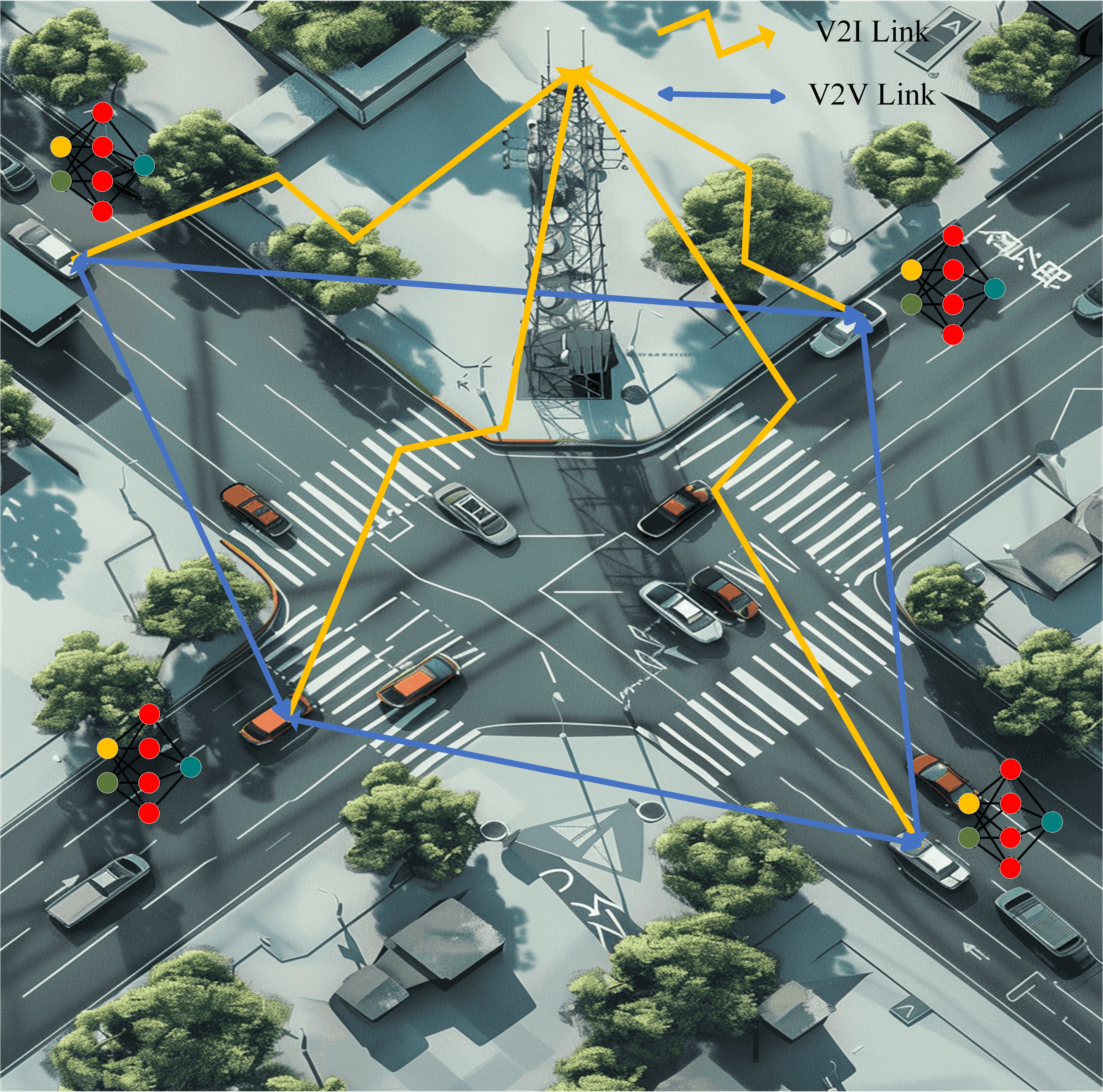}
	\caption{System model.}
	\label{fig1}
	\end{figure}
	\textcolor{red}{In this section, we will sequentially introduce the basic structure of the system, the computation method for interference, and the resource selection to elucidate the system model presented in this paper.}
	
	\subsection{\textcolor{red}{Basic Structure}}	
	\textcolor{red}{This paper primarily focuses on V2I and V2V communications within the V2X paradigm. As shown in Fig. \ref{fig1}, we consider a vehicular traffic model at an intersection, where the Base Station (BS) is strategically located at the center of the intersection. Vehicles enter the roads in a random distribution, with a predefined range of speeds; each vehicle randomly selects a speed and maintains a constant velocity. In this environment, vehicles communicate with the BS via V2I links to transmit high-rate entertainment and lifestyle information. Additionally, vehicles utilize V2V links to communicate with each other, transmitting critical safety information during driving, which requires high reliability.}
	
	\subsection{\textcolor{red}{Interference Calculation Method}}		
	\textcolor{red}{Assume there are $m$ cellular users (CUEs) communicating with the base station, denoted as ${M}=[1,2,3,...,m]$, and $k$ pairs of V2V users (VUEs), denoted as ${K}=[1,2,3,...,k]$. Considering the relatively sparse utilization of uplink resources, to further optimize spectral efficiency, we assume that V2V and V2I links share orthogonally allocated uplink spectrum \cite{10644092}. Therefore, the Signal-to-Interference-plus-Noise Ratio (SINR) of the $i$-th CUE can be expressed as:}
	\begin{equation}
		\gamma^{c}[i]=\frac{P_{i}^{c} h_{i}}{\sigma^{2}+\sum_{j \in K} \rho_{j}[i] P_{j}^{v} \tilde{h}_{j}}
		\label{eq1}
	\end{equation}
	where $\sum_{j \in K} \rho_{j}[i] P_{j}^{v} \tilde{h}_{j}$ represents the interference caused to the $i$-th CUE by VUEs using the same channel. In this context, $\rho_{j}[i] = 1$ indicates that the $j$-th VUE and the $i$-th CUE are utilizing the same channel, whereas $\rho_{j}[i] = 0$ otherwise. $P_{j}^{v}$ denotes the transmit power of the $j$-th VUE, and $\tilde{h}_{j}$ signifies the power gain of the $j$-th VUE. The term $\sigma^{2}$ represents the noise power, while $P_{i}^{c}$ indicates the transmit power of the $i$-th CUE, and $h_{i}$ denotes the channel gain. According to Shannon's formula, the communication capacity of the $i$-th CUE can be expressed as:
	\begin{equation}
		C ^ { c } \left[ i \right] = B \cdot \log ( 1 + \gamma ^ { c } \left[ i \right] )
		\label{eq2}
	\end{equation}
	where $B$ represents the channel bandwidth.
	
	Similarly, the SINR for the $j$-th VUE can be expressed as:
	\begin{equation}
		\gamma ^ { v } \left[ j \right] = \frac { P _ { j } ^ { v } \cdot g _ { j } } { \sigma ^ { 2 } + G _ { V 2 I } + G _ { V 2 V } }
		\label{eq3}
	\end{equation}
	 where:
	 \begin{equation}
	 	G _ { V 2 I } = \sum _ { i \in M } \rho _ { j } \left[ i \right] P _ { i } ^ { c } g _ { i , j }
	 	\label{eq4}
	 \end{equation}
	 represents the interference from V2I links within the same resource block and
	 \begin{equation}
	 	G_{V2V}=\sum_{i\in M}\sum_{j^{\prime}\in K,j\neq j^{\prime}}\rho_j[i]\rho_{j^{\prime}}[i]P_{j^{\prime}}^v g_{j^{\prime},j}^v 
	 	\label{eq5}
	 \end{equation}
	 represents the interference from other V2V links within the same resource block, where $g_{j}$ represents the power gain of the $j$-th VUE, while $g_{i,j}$ and $g_{j^{\prime},j}^v$ denote the interference power gains from the $i$-th CUE and the $j^{\prime}$-th VUE, respectively.
	 The channel capacity of the $j$-th VUE can be expressed as follows:
	 \begin{equation}
	 	C ^ { v } \left[ j \right] = B \cdot \log ( 1 + \gamma ^ { v } \left[ j \right] )
	 	\label{eq6}
	 \end{equation}
	 
	\subsection{\textcolor{red}{Resource Selection Method}}
	
	In this paper, we refine the resource selection for vehicles into channel selection and power level selection.
	\begin{itemize}
	\item \textcolor{red}{For a communication network in the IoV, the efficient utilization of channel resources is a prerequisite for maximizing efficiency. It is crucial to avoid scenarios where some channels are congested while others remain idle. Especially in the context of vehicles autonomously selecting channel resources, vehicles need to consider not only the success of their own information transmission but also whether they are occupying the only viable option for other vehicles in challenging situations. Therefore, It cannot achieve global optimization just by choosing the optimal resources for each vehicle based on its own collected data, the overall benefit is the critical issue.}
	\item \textcolor{red}{The choice of power level often occurs after channel resource selection. Smaller transmission power can reduce interference with other links, but it may lead to transmission failure due to its lower signal to noise and interference ratio over the selected link. Conversely, selecting higher transmission power causes greater interference to other links and increases energy consumption. Hence, choosing an appropriate transmission power level is crucial for the overall system performance.}
	\end{itemize}
	
	\textcolor{red}{Given that the resources of V2I links can usually be directly allocated by the base station, we consider employing a distributed algorithm for the V2X resource allocation task under the condition of predetermined V2I resource allocation. Assume the number of subchannels equals the number of V2I links, denoted as $m$, and the number of power levels in the power list for V2V communication is $n$. Each V2V link has a total of $m \cdot n$ resource selection options. The objective is to minimize the interference to V2I links while satisfying the latency requirements and reliability of V2V links, thereby maximizing the transmission rate of V2I links.} To better evaluate the performance of V2V links, we transform the reliability requirements of V2V links into outage probabilities \cite{7270335}.
	
	 The interaction process between the model and the environment can be described as follows:
	 
	 At a larger time scale, vehicles determine neighbor information based on distance relationships, implicitly constructing a graph. At a smaller time scale, vehicles acquire local observations of the environment and collect information transmitted by neighbors. They aggregate these inputs through a model to obtain low-dimensional features that encapsulate global information. Subsequently, policies learned through Deep Reinforcement Learning (DRL) are used to select channels and transmit powers. \textcolor{red}{It is important to note that while vehicles are participants in the system model, in the algorithm, V2V links serve as nodes in the graph and agents in the DRL framework. The final actions derived from algorithmic decisions are executed by the vehicles themselves.} This decentralized approach leverages local interactions and learned strategies to optimize resource allocation in vehicular networks.
	
	\section{Design of Graph Neural Network Models}
	\label{sec4}
	\subsection{Graph Construction}
	\begin{figure}[t]
		\centering
		\includegraphics[width=\columnwidth]{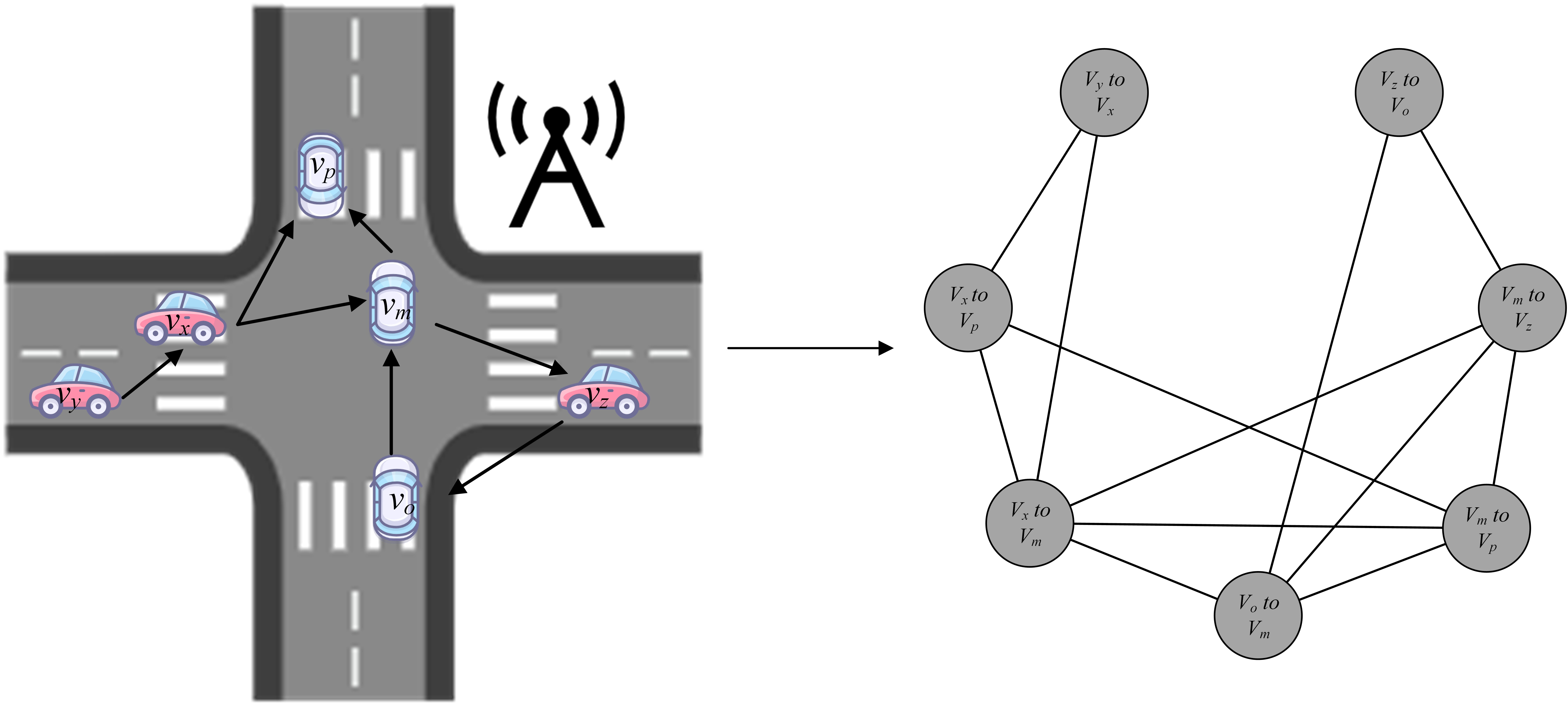}
		\caption{Vehicle network graph construction.}
		\label{fig2}
	\end{figure}
	To construct a graph representation of the vehicular network, we treat each V2V pair as a node, denoted as ${N_g} = [v_1, v_2, v_3, ..., v_k]$, and the interference relationships between links as edges. For node $v$, it contains an initial feature vector $x_v$ and a list $N(v)$ storing the indices of its neighboring nodes. The initial feature of a node encapsulates the local observations of channels and interference information by the vehicle. Based on the assumption that the number of subchannels equals the number of CUEs, denoted as $m$, we record for the $i$-th subchannel the instantaneous channel power gain for V2V links, represented as $G_t[i], i \in M$, the subchannel power gain from the transmitter to the receiver in V2I links, denoted as $H_t[i], i \in M$, and the interference signal strength from the previous time slot, denoted as $I_{t-1}[i], i \in M$. Consequently, the feature of node $v$ can be expressed as
	\begin{equation}
		x_v = {\left\{ G_t | | H_t | | I_{t-1} \right\}}
		\label{eq7}
	\end{equation}
	where $||$ denotes the concatenation of vectors.\par
	Existing methods predominantly construct the vehicular network as a complete graph, emulating the mutual interference relationships among all links. However, when the number of vehicles in the environment is substantial, the graph becomes exceedingly complex, and the computational load for aggregating node features escalates, leading to prolonged decision-making times and potentially introducing additional delays. To address this issue, we propose constructing the graph based on the communication relationships between vehicles. Since each V2V pair consists of a transmitting vehicle and a receiving vehicle, and assuming there are s vehicles in the environment, represented as ${v} = [V_1, V_2, V_3, ..., V_s]$. We assign a unique label to each V2V link, denoted as $V_t to V_r$, where $V_t,V_r \in V$. Here, \( V_t \) denotes the transmitting vehicle, and \( V_r \) denotes the receiving vehicle. \textcolor{red}{Fig. \ref{fig2} illustrates our proposed graph construction method based on communication relationships.} Assume that vehicle \( V_x \) has three destination vehicles: \( V_m \), \( V_p \), and \( V_o \). At the same time, \( V_x \) is also the target vehicle for vehicles \( V_y \) and \( V_z \). The labels for the V2V link nodes formed by these vehicles are as follows: \( V_x \) to \( V_m \), \( V_x \) to \( V_p \), \( V_x \) to \( V_o \), \( V_y \) to \( V_x \), and \( V_z \) to \( V_x \). For each node, taking the node \( V_x \) to \( V_m \) as an example, we consider all nodes with labels containing either \( V_x \) or \( V_m \) as neighbors of the node \( V_x \) to \( V_m \). This includes all nodes labeled as \( V_x \) to \( Z \), \( Z \) to \( V_x \), \( V_m \) to \( Z \), and \( Z \) to \( V_m \), where \( Z \) represents any vehicle that meets the conditions.
	
	We stipulate that each vehicle selects three transmission destinations from a list of nearby neighboring vehicles, meaning that each vehicle initiates three V2V links. As a result, the total number of V2V links in the environment is $3s$, where s is the number of vehicles. For each vehicle, the number of its neighboring vehicles can be approximated to be around 12. This approximation is based on the fact that each vehicle $(V_x,V_m)$ has three outgoing links $(V_x to Z, V_m to Z)$ and an uncertain number of incoming links $(Z to V_x, Z to V_m)$. However, since the total number of links is three times the number of vehicles, the average number of times a vehicle serves as a receiver is also 3. Therefore, the number of neighbors for each vehicle is approximately 12 and remains constant regardless of the increase in the number of vehicles in the environment. In contrast, in a complete graph, the number of neighbors for each node is $3s - 1$, which continuously increases with the number of vehicles, leading to a higher complexity as the number of vehicles grows.
	
	\textcolor{red}{In conclusion, the graph construction method based on communication relationships that we propose can limit the number of neighbors for each node to a relatively stable count. Although it overlooks some interference relationships between links, it essentially preserves the strongest interference from neighbors. When the vehicle density is high, features in the graph can still be extracted with a relatively small computational load. Moreover, due to the characteristic of feature propagation layer-by-layer in the graph, even if there is no direct connection between nodes, they can still be influenced by distant nodes after multiple iterations.} \textcolor{red}{Furthermore, the rationale behind choosing to construct the graph based on communication relationships rather than physical distances lies in two aspects: on one hand, this approach intuitively reflects the communication relationships between links, and the edges between nodes reflect the interference between links. On the other hand, this method reduces the communication requirements for vehicles when collecting data needed for decision-making, thereby facilitating the implicit construction of the graph among vehicles. If the graph is constructed based on the physical distances, the topology of the graph is varying too fast as the vehicles moving, which may cause high complexity for accurately keeping the graph structure, while the variation of physical distance bias in the graph also leads to more bias on the interference estimation of different links.} For the six neighbor nodes represented by $V_x \to Z$ and $V_m \to Z$, vehicles $V_x$ and $V_m$ possess all relevant information. As for the neighbor nodes represented by $Z \to V_x$ and $Z \to V_m$, they can carry node features and other pertinent information when transmitting to $V_x$ and $V_m$. Thus, by establishing a bidirectional connection between $V_x$ and $V_m$, information can be transmitted within the graph without incurring additional communication overhead.
	
	We have performed a comparative analysis of the computational load in graph networks under both complete and non-complete graph scenarios. With $s$ vehicles in the environment, the graph contains $3s$ nodes. Assuming each node has a feature dimension of $D_{input} = 60$ and an output dimension of $D_{output} = 20$, considering a single-layer aggregation model. The number of neighbors for each node is $N_{nei}$, thus, the required multiplication operations can be calculated as follows:
	\begin{equation}
		N = D_{input} \cdot D_{output} \cdot 3s \cdot N_{nei}
		\label{eq8}
	\end{equation}
	
	When constructing a complete graph, $N_{nei} = 3s - 1$, whereas for the graph construction method we propose, $N_{nei} = 12$. Essentially, the difference in the number of multiplication operations between these two graph construction methods is determined by the number of neighbors, and the ratio of their computational loads is also the ratio of the number of neighbors. Therefore, the property of our proposed method, where the number of neighbors does not increase with the number of vehicles, becomes particularly significant.
	
	To better characterize the vehicular network environment, we assign a weight to each edge in the graph. Generally, the interference caused by links that are farther away is smaller to the current link. Therefore, we record the distances between transmitters, represented as a square matrix $D = [d_{11}, d_{12}, ..., d_{1s}; d_{21}, d_{22}, ..., d_{2s}; ...; d_{s1}, d_{s2}, ..., d_{ss}]$. Suppose the interference between the links $V_m$ to $V_n$ and $V_x$ to $V_y$ corresponds to the edge between nodes $v_p$ and $v_q$, the weight of which can be represented as:
	\begin{equation}
		\delta_{pq} = 1 - \frac{d_{mx}}{max(D)}
		\label{eq9}
	\end{equation}
	\subsection{GNN Network}
	After constructing the graph, we will introduce the GNN model. Due to the frequent changes in the number of vehicles and the communication relationships between vehicles in vehicular networks, traditional GNN models struggle to adapt to the rapid changes in node count and adjacency matrix, necessitating frequent adjustments to the graph structure such as pruning or adding edges, and even requiring retraining of the model. Therefore, we introduce the GraphSAGE model, which is adept at handling dynamic and large-scale graph structures, and enhance it by incorporating edge weights into the node feature aggregation process.
	
	The operation of the GraphSAGE model can be divided into three main steps: neighbor sampling, feature aggregation, and feature update. Neighbor sampling is applicable to large-scale graph processing, where the number of neighbors to be sampled for each node at various layers is predefined, allowing for the random selection of these neighbors. This random sampling mechanism constrains the computational demand for feature aggregation at each node, ensuring that the model remains computationally feasible even when confronted with exceedingly intricate graph structures. Although both the neighbor sampling function and our proposed graph construction method serve to manage computational resource expenditure, our approach differs from random sampling across all nodes in a complete graph. In our constructed graph, the neighbors sampled at each layer are confined to a defined scope, and the interference they exert on the current node decreases with the increasing layer depth. This hierarchical reduction in interference facilitates the extraction of superior features, as the influence of second-layer neighbors on the current node is less than that of first-layer neighbors.
	
	Assuming the number of neighbors sampled is $S$, for a node $v$, the set of sampled neighbors is denoted as $N_s(v)$, and the aggregated node feature is represented as $z_v$. The feature aggregation process can be expressed as follows:
	\begin{equation}
		z _ { v } = f _ { a g g r e g a t e } ( \left\{ x _ { u } | u \in N_s ( v ) \right\} )
		\label{eq:10}
	\end{equation}
	where, $f _ { a g g r e g a t e }$ denotes the aggregation function of the model. We employ the mean aggregation function, but with a key distinction: we incorporate edge weights to reflect the importance of each neighboring node's features. This can be represented as:
	\begin{equation}
		z_v = \sigma ( W_a \cdot \sum _ { u \in N_s ( v ) } \frac { X _ { u }\cdot\delta_{uv} } { | N_s ( v ) | } + b_a)
		\label{eq:11}
	\end{equation}
	where, $W_a$ signifies the weights of the aggregation function, $b_a$ denotes the bias term of the aggregation function, and $\sigma$ signifies the activation function.
	
	For the feature update process, we combine the aggregated results of the neighbor features with the initial features of the node itself to obtain the final node feature embedding:
	\begin{equation}
		h _ { v } = f _ { u p d a t e } ( z _ { v } , x _ { v } )
		\label{eq:12}
	\end{equation}
	
	In this approach, we select summation as the method to integrate the aggregated results of the neighbor features with the initial node features. This integration is then processed through feature extraction using weight parameters obtained from training. The resulting node embedding is denoted as $h_v$. Thus, the update function $f_{update}$ can be expressed as follows:
	\begin{equation}
		f _ { u p d a t e } ( z _ { v } , x _ { v } ) = \sigma ( W_u \cdot ( z _ { v } + x _ { v } ) + b_u )
		\label{eq:13}
	\end{equation}
	where, $W_u$ and $b_u$ denote the trained weights and bias vector in the feature update function. Consequently, the complete aggregation process for a single round can be represented as:
	\begin{equation}
		h_v=f _ { u p d a t e } ( x _ { v } , f _ { a g g r e g a t e } ( \left\{ x _ { u } | u \in N_s ( v ) \right\} ) )
		\label{eq:14}
	\end{equation}
	
	Similarly, the two-layer aggregation process can be expressed as:
	\begin{equation}
		\begin{aligned}
			h_{v} &= f_{update}(x_{v}, f_{aggregate}(\{f_{update}(x_{u}, \\
			&\quad f_{aggregate}(\{x_{n} | n \in N_s(u)\})) | u \in N_s(v) \}))
		\end{aligned}
		\label{eq:15}
	\end{equation}
	
	Algorithm \ref{al1} summarizes the feature extraction process of the GraphSAGE model.
				
			%
				%
				%
			%
	\begin{algorithm}[t]
		\SetAlgoLined
		\caption{GraphSAGE}
		\label{al1}
		\textbf{Input:} Graph network model, Node features\;
		\textbf{Output:} Aggregation result\;
		\textbf{Define:} \textcolor{red}{Maximum iteration count $max\_iter$ and convergence threshold $\epsilon$}\;
		\textbf{Initialize:} Initialize model, \textcolor{red}{$iter \gets 0$, $converged \gets False$}\;
		\textbf{\textcolor{red}{Large Time Scale:}} \textcolor{red}{Construct global graph}\;
		\textbf{\textcolor{red}{Small Time Scale:}} \textcolor{red}{Update and Aggregate Node Features}\;
		
		\While{\textcolor{red}{$iter < max\_iter$ \textbf{and} $converged = False$}}{
			\For{each node $v$ from $N_g$}{
				\textbf{Node Neighbor Sampling:} 
				
				Randomly select five neighbors of node $v$ and store them in $N_s(v)$
				
				\For{each neighbor $u \in N_s(v)$}{
					Randomly select five neighbors of node $u$ and store them in $N_s(u)$
				}
				
				\textbf{Feature Aggregation and Update:}
				
				\For{each neighbor $u \in N_s(v)$}{
					\tcp{\textcolor{red}{Update feature of neighbor $u$}}
					$ h_u = f_{update}(x_u, f_{aggregate}(x_n, n \in N_s(u)))$
				}
				
				\tcp{\textcolor{red}{Update feature of node $v$}}
				$ h_v = f_{update}(x_v, f_{aggregate}(h_u, u \in N_s(v)))$
			}
			\tcp{\textcolor{red}{Check for convergence}}
			\textcolor{red}{$converged \gets \text{check\_convergence}()$}
			\textcolor{red}{$iter \gets iter + 1$}
		}
	\end{algorithm}
	
	To ensure that the features extracted by the GraphSAGE model are beneficial for the agent's decision-making, we have designed a unique updating mechanism for it. Since the GraphSAGE model is deployed globally, while the DDQN model only affects a single agent, it is challenging to integrate the two networks for joint training. Therefore, we opt to train the GraphSAGE model separately.
	
	The training of GNNs typically falls into supervised learning, semi-supervised learning, or unsupervised learning. Unsupervised learning can incur significant computational overhead during the training phase, while supervised and semi-supervised learning are limited by the cost of obtaining node labels. Consequently, we aim to find a cost-effective method that provides a learning direction for the network and enhances the decision-making capabilities of the DDQN model.
	
	In fact, we employ the key role of GraphSAGE model to aggregate node features, aiding the agent in discerning the quality of each channel. The rewards obtained after the DDQN network taking actions provide a sound assessment of the channels. Consequently, it can store the reward information acquired by the agent for each sub-channel in a matrix corresponding to the number of sub-channels, $R_g = [ r_1, r_2, r_3, ..., r_m ] $. This matrix serves as the labels for the corresponding nodes.
	
	However, the stored reward values are with a certain latency, making them unsuitable as absolute labels currently. Instead, they still provide a fuzzy directional guide for the network's learning in complex environments. To address such a problem and trying to efficiently use the restored reward values, we propose a more moderate update strategy. This strategy involves using a lagged network to aggregate node features and weighting them with the labels according to a specific ratio. This approach has two main benefits: 
	
	\begin{itemize}
	\item[1)] It weakens the absolute influence of the labels, maintaining the stability of learning.
	\item[2)] It gradually guides the network's predictions towards the ideal label values.	
	\end{itemize}
	
	By incorporating this moderated update strategy, one can ensure a balance between learning stability and directional guidance, ultimately enhancing the performance of the DDQN model in making decisions based on the aggregated features from the GraphSAGE model.	
	
	Specifically, the Mean Squared Error (MSE) function is used here as the loss function for the network, defined as follows:
	\begin{equation}
	L o s s ( \theta ) = \sum _ { v \in N _ { g } } ( y _ { v } - h _ { v } ) ^ { 2 }
	\label{eq:16}
	\end{equation}
	where:
	\begin{equation}
	y _ { v } = \kappa  h^{old} _ { v } + ( 1 - \kappa ) R _ { g } ^ { v }
	\label{eq:17}
	\end{equation}
	where $\theta$ represents the weighting parameters in the GraphSAGE model, $y_v$ denotes the smoothed labels used for network updates, $\kappa$ indicates the weighting factor for the aggregation results, $R_g^v$ represents the label for node $v$, and $h^{old}_v$ signifies the aggregation result from the lagged network.
	
	\section{The GNN-DDQN Model for Resource Allocation Problems}
	\label{sec5}
	In this section, we primarily introduce the method for solving the resource allocation problem in V2X communication based on the GNN-DDQN model. First, we present the key equations of reinforcement learning and the core principles of the DDQN model. Then, we discuss how to combine the DDQN model with the previously mentioned GraphSAGE model to address the resource allocation problem.
	\subsection{Key Formulas in RL and DDQN }
	Reinforcement learning are typically formulated as Markov Decision Processes (MDPs), which manifest as interactions between an agent and an environment. The agent acquires state information from the environment, makes decisions according to a policy $\pi$, and then executes an action. The environment, in turn, provides the agent with a reward and updates the current state.

	Specifically, at each time step $t$, the agent obtains the current state $s_t$ from the state space of the environment and selects an appropriate action $a_t$ from the action space $A$ according to the existing policy $\pi$ to choose the transmission channel and transmission power for the V2V link. The policy $\pi$ is determined by a state-action function, also known as the Q-function, denoted as $ Q(s_t, a_t)$. In deep reinforcement learning, the Q-function is typically approximated using deep learning to adapt to complex environmental changes. After the agent takes an action, the environment will transit to a new state $s_{t+1}$ and provides a reward $r_t$ that evaluates the quality of the action based on the results of the agent's choice. In this paper, the reward is derived from the outage probability of the V2V link and the rate of the V2I link.
	
	Next, we will sequentially introduce the details of state, action, and reward in the DDQN network:
	\subsubsection{State space}
	For the V2X environment we consider, the true state information primarily consists of the vehicle's observations of the environment $x_v$ and the low-dimensional features $h_v$ extracted from these observations by the GraphSAGE model. To assist the agent in making better decisions, each vehicle sends its channel selection information to its target vehicles. Based on this, we collect each agent's neighboring agents' previous channel selection information $N_{t-1}$ and calculate the ratio $L_t$ between the remaining bits to be sent by the vehicle and the total bits that need to be sent, as well as the remaining transmission time $U_t$ under the delay constraint. Combining these pieces of information, the state acquired by the agent is represented as follows:
	\begin{equation}
		S _ { v }^t = \left\{ h _ { v } ^ { t} | | x_v^t | | N _ { t - 1 }^v | | L _ { t }^v | | U _ { t }^v \right\}
		\label{eq:18}
	\end{equation}
	\subsubsection{Action space}
	Based on the collected and observed state information, the DDQN network selects an action $a_t \in A$ according to the policy $\pi$. Since the agent needs to select both the subchannel and the power level simultaneously, we combine these two types of actions into a composite action. The composite action selected by the agent is mapped onto two dimensions, representing the choices of subchannel and power level, respectively. In this paper, we consider a relatively simple case with three power levels and $m$ resource blocks. Therefore, there are a total of $3 \times m$ possible actions. Assuming the agent selects action $a_t$, we decompose it as follows:
	\begin{equation}
		a _ { r } ^ { t } = a _ { t } \% m
		\label{eq:19}
	\end{equation}
	and
	\begin{equation}
		a _ { p } ^ { t } = a _ { t } / m
		\label{eq:20}
	\end{equation}
	where $\%$ denotes the modulo operation, and $/$ represents division with rounding down to the nearest integer. $a_r^t$ denotes the decomposed subchannel selection action, and $a_p^t$ represents the choice of power level.
	\subsubsection{Reward function}
	The objective of reinforcement learning is to enable V2V links in the environment to meet the low latency and high reliability requirements of V2V communication to the maximum extent, while minimizing interference to V2I links to maximize the transmission rate of V2I links. Previously, we transformed the latency and reliability requirements of V2V links into a requirement for outage probability. Therefore, for the design of the reward, we only need to consider the outage probability of V2V links and the transmission rate of V2I links. Additionally, we have set a penalty term based on the time already spent on the link transmission. The reward function is expressed as follows:
	\begin{equation}
		\begin{aligned}
			r_{t} &= \lambda_{c} \sum_{i \in M} C^{c}[i] + (1 - \lambda_{c}) \sum_{j \in K} C^{v}[j] \\
			&\quad - \lambda_{p} (T_{0} - U_{t})
		\end{aligned}
		\label{eq:21}
	\end{equation}
	where, $ \lambda_c $ represents the weight of the V2I link, and $ \lambda_p $ represents the weight of the utilized transmission time. $ U_t $ denotes the remaining time, and $ T_0 $ signifies the transmission delay limit. Therefore, $(T_0 - U_t)$ indicates the time used for transmission. The agent's long-term discounted return can be expressed as:
	\begin{equation}
		R _ { t } = E \left[ \sum _ { n = 0 } ^ { \infty } \beta ^ { n } r _ { t + n } \right]
		\label{eq:22}
	\end{equation}
	where, $ \beta \in [0, 1] $ represents the discount factor for the reward. A larger $ \beta $ indicates that the agent has a longer-term perspective, while a smaller $ \beta $ means the agent is more focused on immediate rewards.
	
	In deep reinforcement learning, the primary objective is to learn an optimal policy $\pi^*$ that maximizes the long-term discounted reward, as well as to develop a deep network model that can predict the Q-values corresponding to state-action pairs. For a given state-action pair $(s_t, a_t)$, the Q-value $Q(s_t, a_t)$ represents the expected cumulative discounted reward obtained after taking action $a_t \in A$ according to policy $\pi$. Thus, the Q-value can be used to assess the quality of an action in a given state. Once we have an accurate estimation of the Q-values, actions can be selected according to the following equation:
	\begin{equation}
		a _ { t } = a r g \max_{a\in A} Q ( s _ { t } , a )
		\label{eq:23}
	\end{equation}
	
	This implies that we select the action with the maximum Q-value. In the DDQN, the Q-values corresponding to the optimal policy, denoted as $Q^*$, can be obtained through the following update equation:
	\begin{equation}
		\begin{aligned}
			Q_{new}(s_{t}, a_{t}) &= Q_{old}(s_{t}, a_{t}) + \alpha [ r_{t+1} +  \\
			&\quad \beta Q_{old}^{target}(s_{t+1}, \arg\max_{a \in A} Q_{old}(s_{t+1}, a)) \\
			&\quad - Q_{old}(s_{t}, a_{t}) \Big]
		\end{aligned}
		\label{eq:24}
	\end{equation}
	where, $\alpha$ is the learning rate, and the second term on the right-hand side of the equation is the Temporal Difference (TD) error used to update the Q-value. The discount factor is denoted by $ \gamma $. $ Q_{\text{old}} $ represents the Q-value predicted by the Q-network, while $ Q_{\text{old}}^{\text{target}} $ represents the Q-value predicted by the target Q-network.
	
	Assuming the weights of the Q-network are denoted by $\varphi$, the output will be the Q-values corresponding to each action when the input is the observed state of the agent. We optimize the parameters $ \varphi$ of the Q-network using the TD error, which can be expressed as follows:
	\begin{equation}
		L o s s ( \varphi ) = \sum _ { ( s_t , a _ { t } ) \in D } ( y _ { Q } - Q ( s _ { t } , a _ { t } , \varphi ) ) ^ { 2 }
		\label{eq:25}
	\end{equation}
	where, ${D}$ denotes the set of state-action pairs, and $ y_Q $ represents the expected target Q-value, which can be given by the following equation:
	\begin{equation}
		y _ { Q } = r _ { t } + \beta Q _ { o l d } ^ { t a r g e t } ( s _ { t + 1 } , a r g \max_{a\in A} Q _ { o l d } ( s _ { t + 1 } , a ) , \varphi ')
		\label{eq:26}
	\end{equation}
	where, $\varphi'$ denotes the parameters of the target Q-network.
	
	\subsection{GNN-DDQN Framework and Training-Testing Process}
\begin{figure}[t]
		\centering
		\includegraphics[trim=10pt 10pt 10pt 10pt, clip, width=\columnwidth]{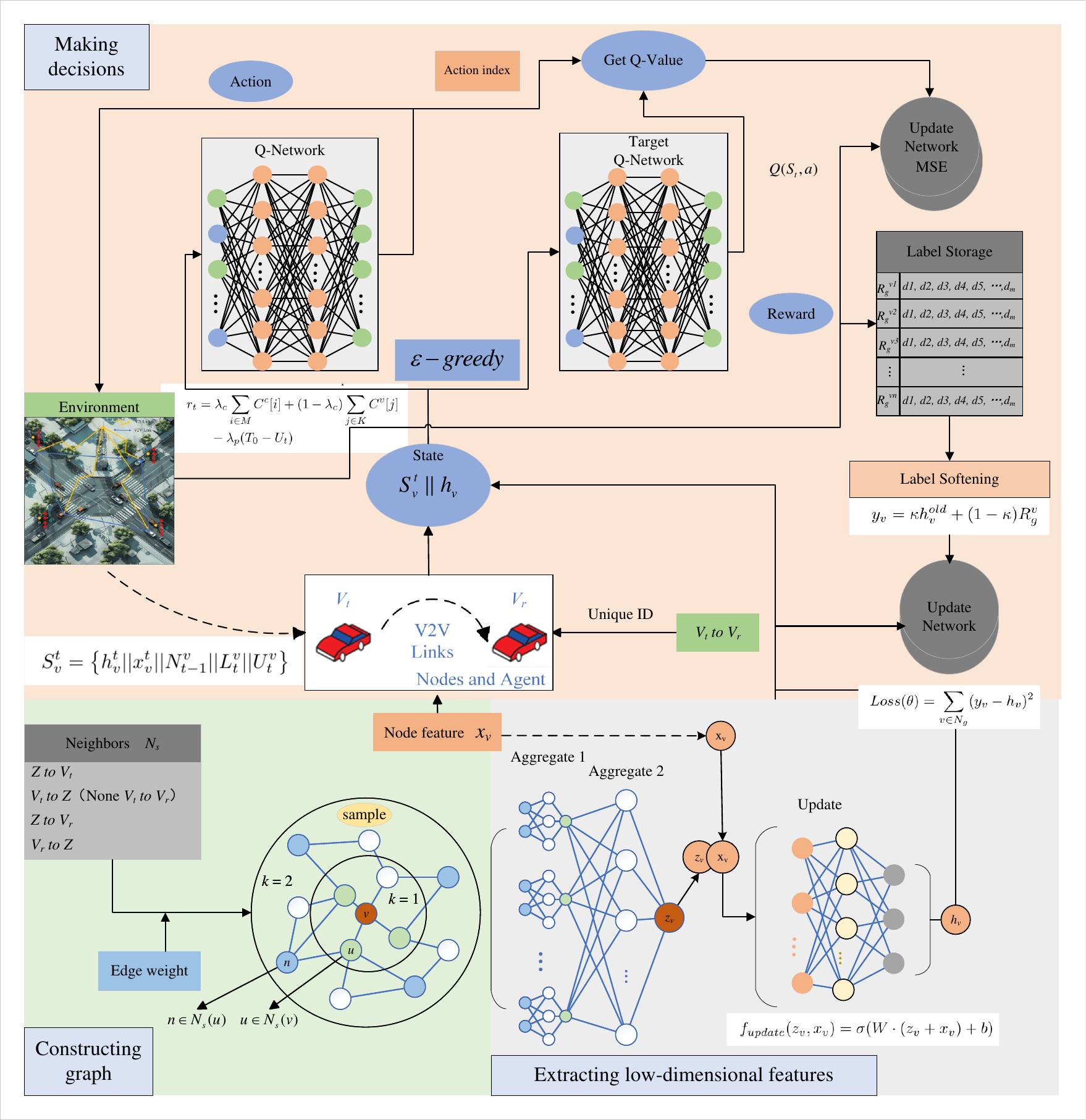}
		\caption{GNN-DDQN framework.}
		\label{fig3}
	\end{figure}

	\textcolor{red}{Fig. \ref{fig3} provides a detailed illustration of the algorithm's structure and operational steps.} The operation of the GNN-DDQN model can be divided into three steps: constructing the graph, extracting low-dimensional features about global information, and making decisions based on the Q-values predicted by the Q-network. Note that the training process of the GraphSAGE model can be referenced from the previously introduced content. Here, we will focus on the training of the DDQN model and the testing of the GNN-DDQN model.
	
	In reinforcement learning, all data required by the agents are generated through interactions between the agents and the environment. We use the tuple $(s_t, a_t, r_t, s_{t+1})$ as a single training sample and store it in a replay buffer, from which samples are periodically drawn to update the DDQN network.
	The environment simulator comprises VUEs, CUEs, and their channel models. Vehicles in the environment are randomly placed at the intersection model according to a uniform distribution. Based on the distance relationships between vehicles, V2V and V2I communication links are established. The agent selects the channel and transmission power based on the CSI in the environment. The environment then updates the fading information and interference information accordingly, resulting in a new state $s_{t+1}$. Concurrently, the agent receives a reward $r_t$ based on the quality of the action taken.
	During the training phase, we adopt an $\epsilon$-greedy policy to balance exploration and exploitation. In the testing phase, actions are selected based on the predicted Q-values, specifically choosing the action with the highest Q-value.
	
	Upon initialization, the agent, operating on a large temporal scale, identifies its neighbors based on vehicle-to-vehicle communication links, implicitly constructing a graph where edge weights are determined by proximity. On a smaller temporal scale, the agent collects local observations denoted as $S_v^t$ and aggregates these with features from neighboring agents to perform feature updates, yielding a low-dimensional representation $h_v$. This representation, combined with the local observations $S_v^t$, forms the agent's state. The agent then predicts the action with the highest Q-value using the Q-network. Rewards are obtained from the environment. During training, the Q-values predicted for each action are stored as baseline evaluations, which are then softened to yield $y_v$, guiding the updates of the GNN model. The agent's strategy is refined using Q-values predicted by the target Q-network, which also updates the Q-network itself. In the testing phase, the action with the maximum predicted Q-value from the Q-network is directly executed as the decision.
	
	\begin{algorithm}[t] 
		\caption{Training Process for GNN-DDQN} 
		\label{al2}
		\KwIn{GraphSAGE model, Q-network model, Simulation environment} 
		\KwOut{Q-network, GraphSAGE-network} 
		\textbf{Define:} \textcolor{red}{Maximum iteration count $max\_iter$ and convergence threshold $\epsilon$}\;
		\textbf{Initialize:} Randomly initialize policy $\pi$, initialize model, start environment simulation and add vehicles, CUEs, and VUEs, \textcolor{red}{$iter \gets 0$, $converged \gets False$}\;
		\textbf{Large Time Scale:} Construct global graph\;
		\textbf{Small Time Scale:} Joint GNN-DDQN model selects actions\;
		
		\While{\textcolor{red}{$iter < max\_iter$ \textbf{and} $converged = False$}}{
			\tcp{\textcolor{red}{Execute feature extraction and aggregation}}
			\textbf{Execute Algorithm \ref{al1}}
			
			\tcp{\textcolor{red}{Run the DDQN Network}}
			For node $v$, the agent selects subchannels and transmission power based on policy $\pi$ using $h_v$ (\textcolor{red}{Eq. \ref{eq:18}}) and locally observed states
			
			The environment simulator generates a new state $s_{t+1}$ and reward $r_t$ based on the agent's actions (\textcolor{red}{Eq. \ref{eq:21}})
			
			Store the selected subchannels and corresponding rewards in $R_g$
			
			Collect and store the tuple $(s_t, a_t, r_t, s_{t+1})$ in the experience replay buffer
			
			Sample a mini-batch from the experience replay buffer
			
			Train the DDQN network using the mini-batch (\textcolor{red}{Eq. \ref{eq:24} and Eq. \ref{eq:25}})
			
			Update the GNN network using labels derived from $R_g$
			
			\tcp{\textcolor{red}{Update policy $\pi$}}
			Update policy $\pi$: select the action with the highest Q-value (\textcolor{red}{Eq. \ref{eq:23}})
			
			\textcolor{red}{$converged \gets \text{check\_convergence}()$}
			\textcolor{red}{$iter \gets iter + 1$}
		}
	\end{algorithm}
	
	\begin{algorithm}[t]
		\caption{Testing Process for GNN-DDQN} 
		\label{al3}
		\KwIn{Trained Q-network model, Trained GraphSAGE model, Simulation environment} 
		\KwOut{Evaluation results} 
		\textbf{Define:} \textcolor{red}{Maximum iteration count $max\_iter$}\;
		\textbf{Initialize:} Load the Q-network model and GraphSAGE model, start the environment simulation, and add vehicles, CUEs, and VUEs, \textcolor{red}{$iter \gets 0$}\;
		\textbf{Large Time Scale:} Construct global graph\;
		\textbf{Small Time Scale:} Joint GNN-DDQN model selects actions\;
		
		\While{$iter < max\_iter$}{
			\tcp{\textcolor{red}{Execute feature extraction and aggregation}}
			\textbf{Execute Algorithm \ref{al1}}
			
			\tcp{\textcolor{red}{Run the DDQN Network}}
			For node $v$, the agent selects the action with the highest Q-value based on $h_v$ (\textcolor{red}{Eq. \ref{eq:18}}) and locally observed states
			
			The environment simulator updates the environment based on the agent's actions (\textcolor{red}{Eq. \ref{eq:21}})
			
			Update the evaluation results, which include the average V2I capacity and the success rate of V2V communication
			
			\textcolor{red}{$iter \gets iter + 1$}
		}
	\end{algorithm}
	
	Algorithm \ref{al2} explicitly expresses the training process of the GNN-DDQN model, while Algorithm \ref{al3} illustrates the testing process of the GNN-DDQN model.
	\section{Simulation Results and Analysis}
	\label{sec6}
				
	\subsection{Simulation settings}
	The simulations in this study were conducted using Python 3.8 and TensorFlow 2.6, adopting settings similar to those assumed in the literature \cite{8633948}. We consider a single-cell system operating at a 2 GHz carrier frequency. The simulations follow the Manhattan scenario setup as described in 3GPP TR 36.885 \cite{3rdGP}, covering both Line of Sight (LOS) and Non-Line of Sight (NLOS) channel conditions. The distribution of vehicles is modeled as a spatial Poisson process, randomly distributed across each lane. It assumes that each vehicle established V2V communication links with its three nearest neighbors, resulting in the number of V2V links being three times the number of vehicles.
	
	We utilize a GraphSAGE model with a depth of 2, sampling 5 neighbors at each layer, and incorporating the node's own features for updates. The feature dimension extracted by the graph network set as 20. Each node's feature dimension input to the graph network is 60, and an average aggregation function is adopted. For the DDQN model, the state dimension input is 102, employing a three-layer neural network model with 500, 250, and 120 neurons in each layer, respectively. The final output the Q-values for 60-dimensional actions. A Rectified Linear Unit (ReLU) function is used as the nonlinear activation function between layers. The activation function can be expressed as follows:
	\begin{equation}
	f _ { R e l u } ( x ) = m a x ( 0 , x )
	\label{eq:27}
	\end{equation}
	
	The learning rate is set to decrease over the training period. The initial learning rate for the graph network is selected as 0.01, and for the DDQN network, it is selected as 0.005, with both having a minimum value of 0.0001. More detailed parameter settings are provided in Table \ref{tab:system_parameters}.
	\begin{table}[t]
		\caption{System Parameters}
		\label{tab:system_parameters}
		\footnotesize
		\centering
		\begin{tabular}{|c|c|}
			\hline
			\multicolumn{2}{|c|}{Parameters of System Model} \\
			\hline
			\textbf{Description} & \textbf{Specification} \\
			\hline
			Carrier Frequency & 2 GHz \\
			\hline
			Height of Vehicle Antenna & 1.5 meters \\
			\hline
			Bandwidth of Single Subchannel & 1.5 MHz \\
			\hline
			Gain of Vehicle Antenna & 3 dBi \\
			\hline
			Height of Base Station Antenna & 25 meters \\
			\hline
			Noise Figure of Vehicle Receiver & 9 dB \\
			\hline
			Gain of Base Station Antenna & 8 dBi \\
			\hline
			Vehicle Speed & 36 km/h to 54 km/h \\
			\hline
			Noise Figure of Base Station Receiver & 5 dB \\
			\hline
			Distance Threshold for Neighbor Vehicles & 150 meters \\
			\hline
			Number of Lanes in Environment & 4 per direction, total 16 lanes \\
			\hline
			Maximum Delay for V2V Link & 100 ms \\
			\hline
			Transmission Power Levels & [23, 10, 5] dBm \\
			\hline
			Noise Power & -114 dBm \\
			\hline
			Weight Coefficients [$\lambda_c$, $\lambda_p$] & [0.3, 1] \\
			\hline
			First Layer Neighbor Sampling Count & 5 \\
			\hline
			Second Layer Neighbor Sampling Count & 5 \\
			\hline
			Depth of GraphSAGE & 2 \\
			\hline
			Dimension of Node Feature Output & 20 \\
			\hline
		\end{tabular}
		\begin{tablenotes}
			\footnotesize
			\item Note: Parameters such as "Carrier Frequency", "Height of Vehicle Antenna", "Bandwidth of Single Subchannel", "Gain of Vehicle Antenna", "Height of Base Station Antenna", and others are adapted from \cite{8633948}.
		\end{tablenotes}
	\end{table}
	
	\begin{figure}[h]
		\centering
		\includegraphics[width=\columnwidth]{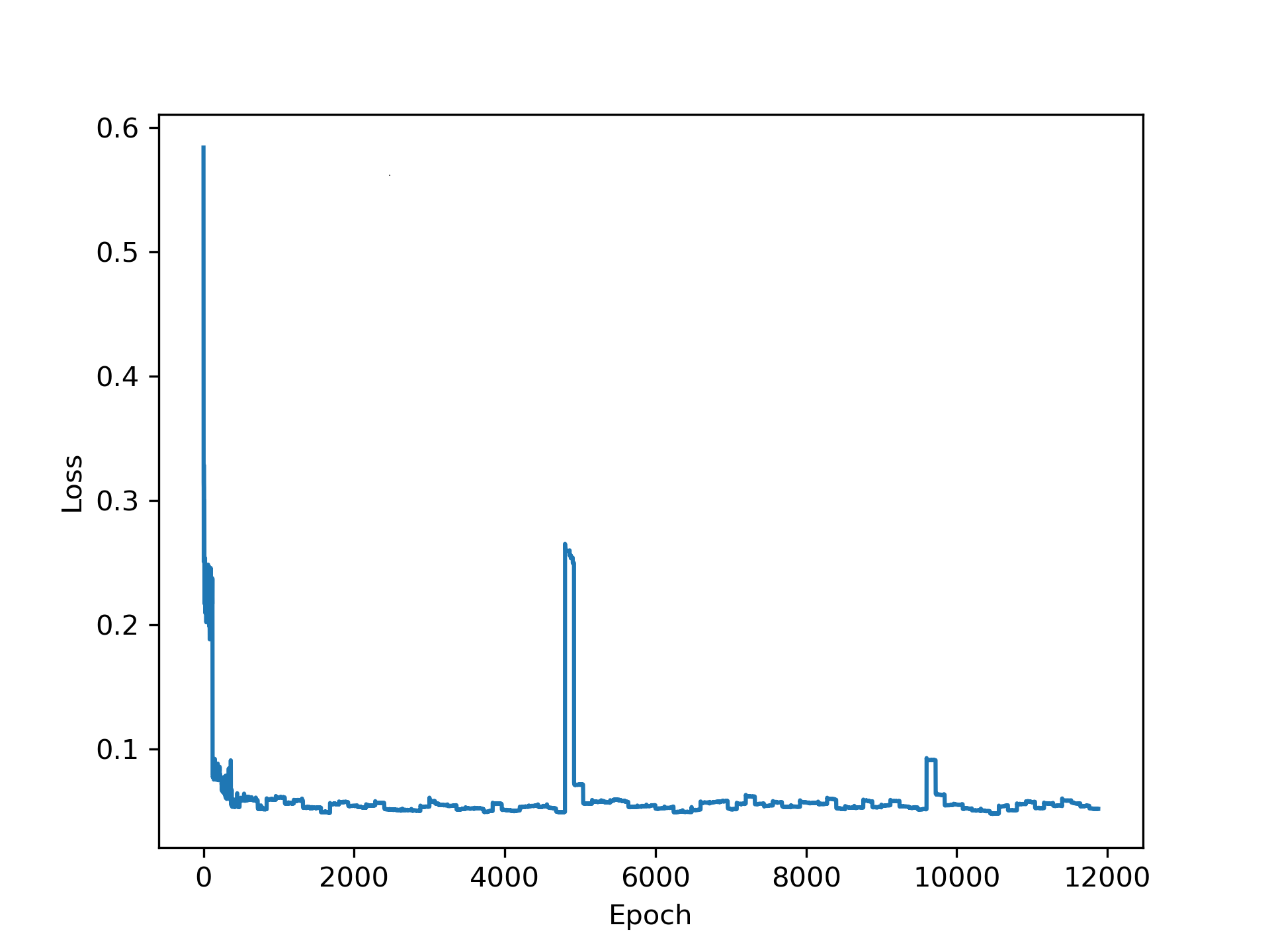}
		\caption{Training loss of GraphSAGE.}
		\label{fig4}
	\end{figure}
	
	\begin{figure}[h]
		\centering
		\includegraphics[width=\columnwidth]{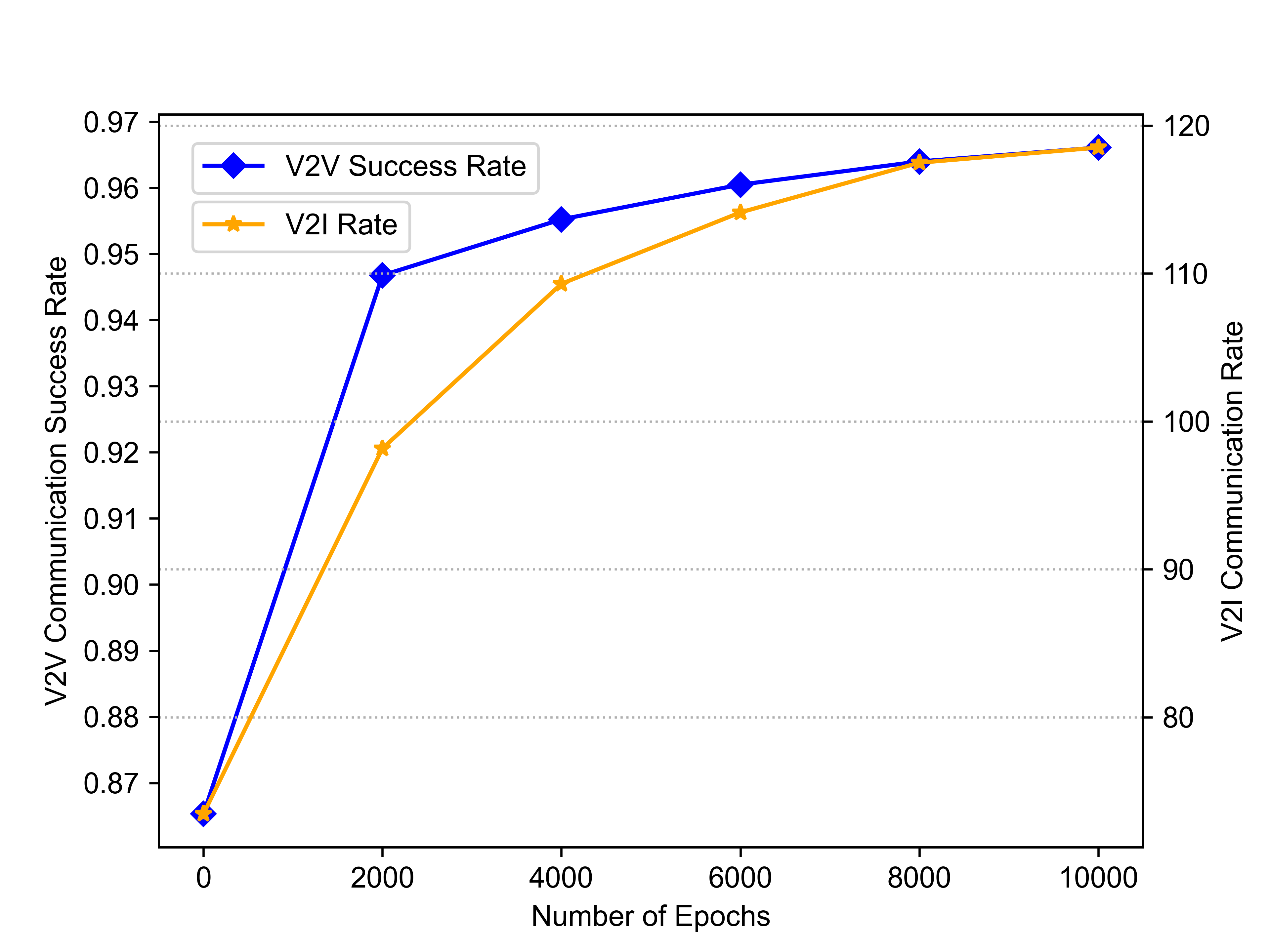}
		\caption{Training effect of GNN-DDQN.}
		\label{fig5}
	\end{figure}
	
	\subsection{Model Training Status}
	To analyze the training process of the network, we monitored the training progress. Since our GraphSAGE model and DDQN network are trained separately, we will analyze the training performance of the two networks individually.
	
	\begin{itemize}
		\item[1)] Training Loss of the GraphSAGE Model: Fig. \ref{fig4} illustrates the loss performance of the GraphSAGE model over 5000 training iterations, where the GraphSAGE model is updated 120 times every 50 iterations. It can be observed that the loss converges rapidly. Since the environment is reset every 2000 iterations, there is a significant fluctuation in loss whenever the environment is reset. However, as the number of iterations increases, the fluctuation due to environmental resets diminishes, and the model begins to adapt to different environments.
		
		\item[2)] \textcolor{red}{Performance Comparison of the GNN-DDQN Network at Different Training Iterations: Fig. \ref{fig5} depicts the testing performance of the policy obtained after 10,000 training iterations in the simulated environment. In practice, in order to achieve a convergent policy, we conducted nearly 40,000 training iterations. Here, we present the results of the first 10,000 training iterations. It can be observed that as the number of training iterations increases, the evaluation V2I communication rate and average V2V communication success rate in the model tests gradually increase, although the rate of increase gradually slows down.}
	\end{itemize}
	
	\begin{figure}[t]
		\centering
		\includegraphics[width=\columnwidth]{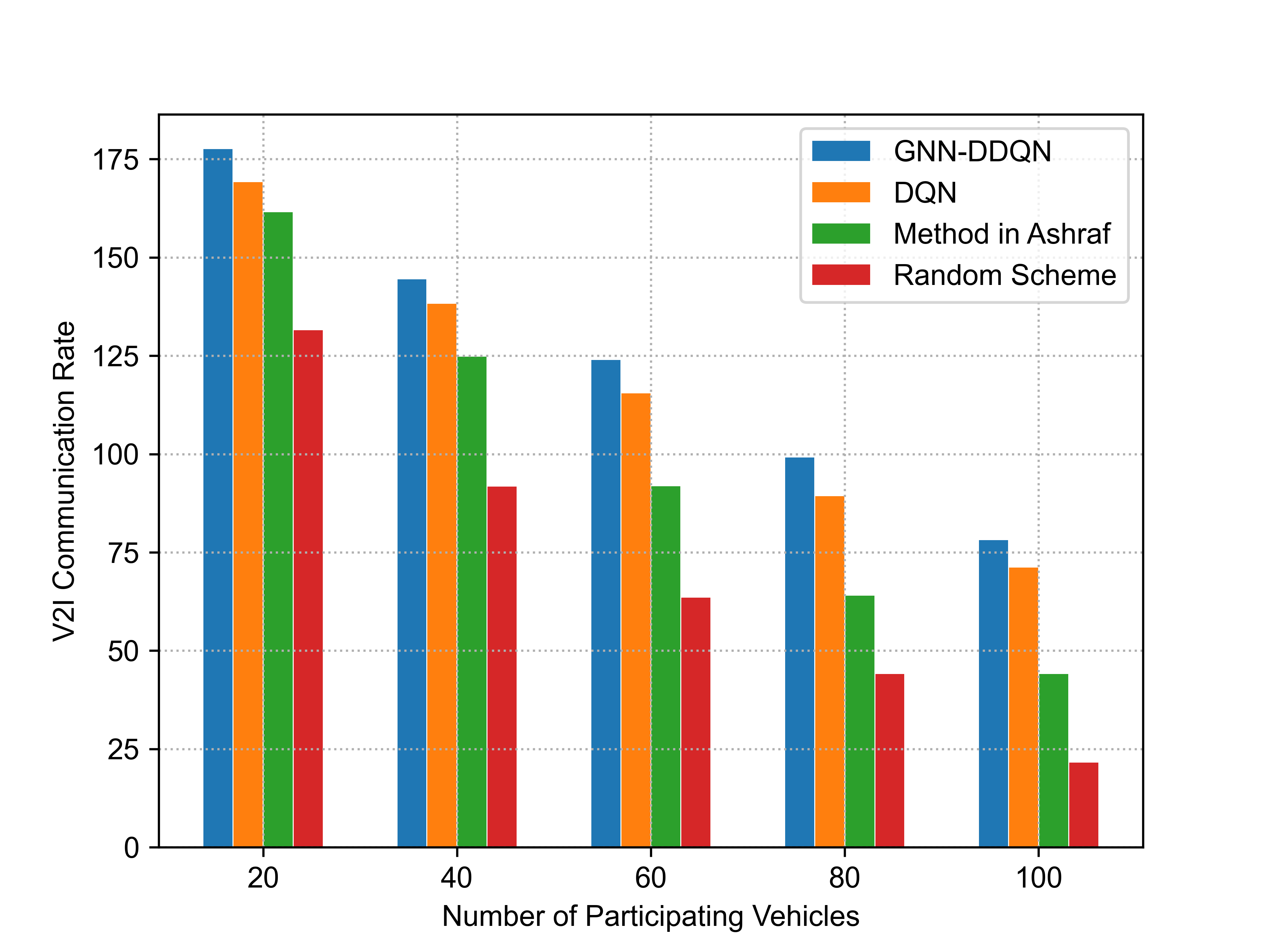}
		\caption{Relationship between average V2I rate and number of vehicles.}
		\label{fig6}
	\end{figure}
	
	\subsection{Static Environment Testing}
	Three baseline methods are used for comparison. The first is the random resource allocation method, where the agent randomly selects channels and power levels, serving as a lower bound. The second is the one mentioned in \cite{7848885}, which groups vehicles with similar channel conditions together and iteratively allocates sub-channels. The third is the one used in \cite{8633948}, which employs a general DQN model for resource allocation, helping us analyze the performance gains brought by introducing GNN.
	
	We divide the decision-making process of all agents into ten batches to address the issue of resource collisions caused by information delay when agents simultaneously select channels. Once the agents have made their decisions, they keep their actions fixed until the next round of reallocation. To obtain more generalizable policies, we periodically reset the environment during the training phase.
	
	When obtaining test results and plotting graphs, we reset the environment 100 times, taking the average of 200 samples for each environment, and then averaging the data across the 100 environments. We observed that the policies learned by the agents exhibit fluctuations in changing environments, but the fluctuation range is controllable. Given that factors such as vehicle density can vary significantly across different environments, there is no constant optimal solution, which makes the presence of fluctuations reasonable.
	
	\begin{figure}[t]
		\centering
		\includegraphics[width=\columnwidth]{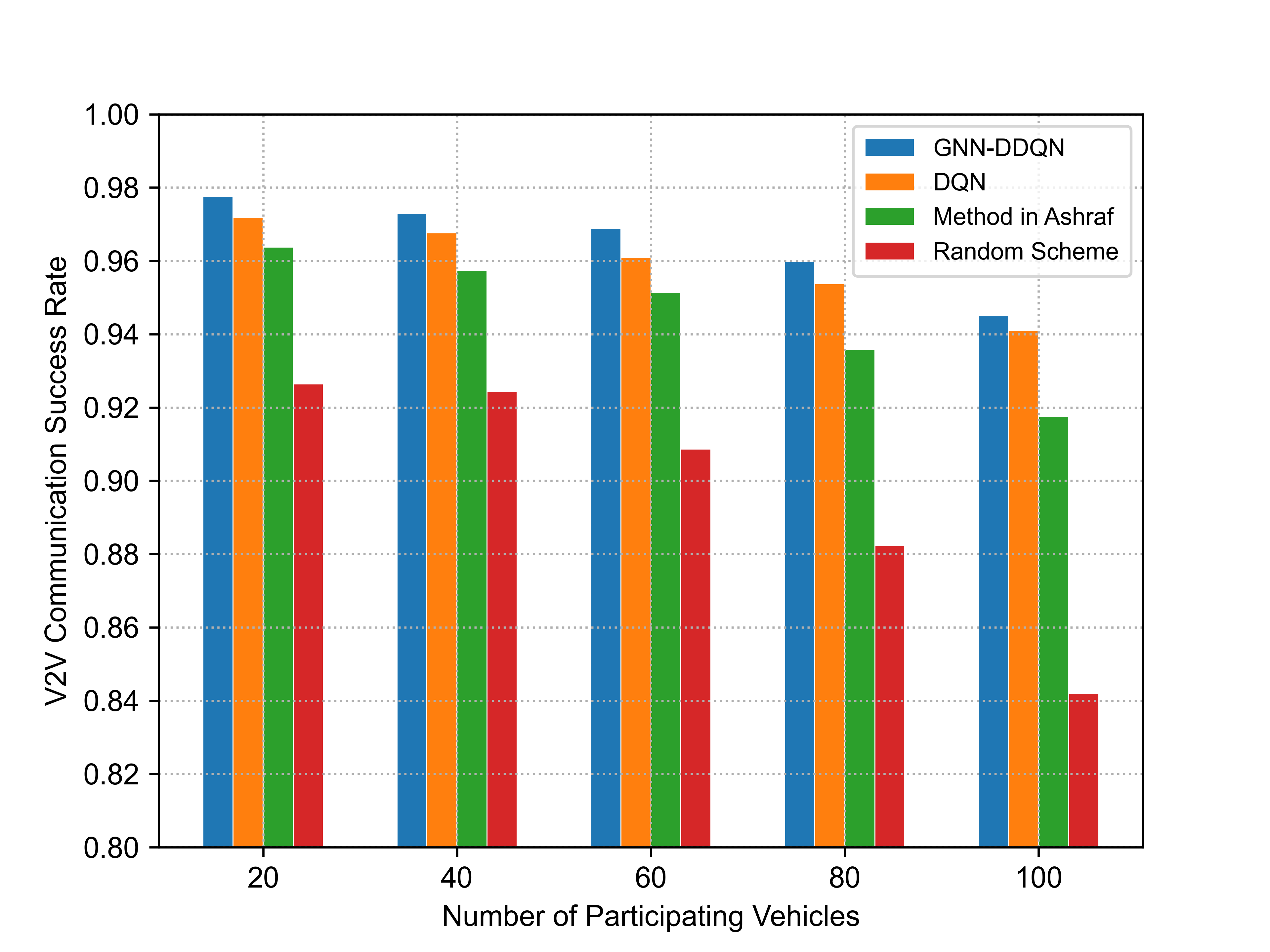}
		\caption{Relationship between average V2V success rate and number of vehicles.}
		\label{fig7}
	\end{figure}
	
	\begin{itemize}
		\item[1)] V2I Communication Rate: Fig. \ref{fig6} illustrates the relationship between the total V2I rate across all sub-channels and the number of vehicles. From the figure, it is evident that as the number of vehicles increases, the number of V2V links in the environment also increases. Consequently, the interference from V2V communication links to V2I links increases, resulting in a decrease in the V2I link rate. Although both the DQN method and our proposed GNN-DDQN method are fundamentally based on reinforcement learning, the incorporation of GNN allows the agent to acquire richer and more comprehensive global state information, thereby leading to superior performance. \textcolor{red}{Strategies based on traditional methods and random strategies suffer particularly severe interference, as reinforcement learning methods can learn policies that maximize long-term rewards and thus perform better in high-density communication environments.}
		
		\item[2)] V2V Communication Success Rate: Fig. \ref{fig7} illustrates the relationship between the transmission success rate of all V2V links in the environment and the number of vehicles. As previously mentioned, the number of V2V links in the environment is three times the number of vehicles. \textcolor{red}{It can be observed that as the number of vehicles increases, the overall communication success rate shows a declining trend, which is consistent with common sense. The success rate of V2V communication using a random strategy declines at the fastest rate. Other methods exhibit a slower rate of decline, indicating their capabilities of improved resource utilization. Evidently, our proposed GNN-DDQN method outperforms the conventional DQN network, further demonstrating that GNN can assist DRL in enhancing the system performance.}
	\end{itemize}
	
	\subsection{Strategy Analysis}
\begin{figure}[t]
		\centering
		\includegraphics[width=\columnwidth]{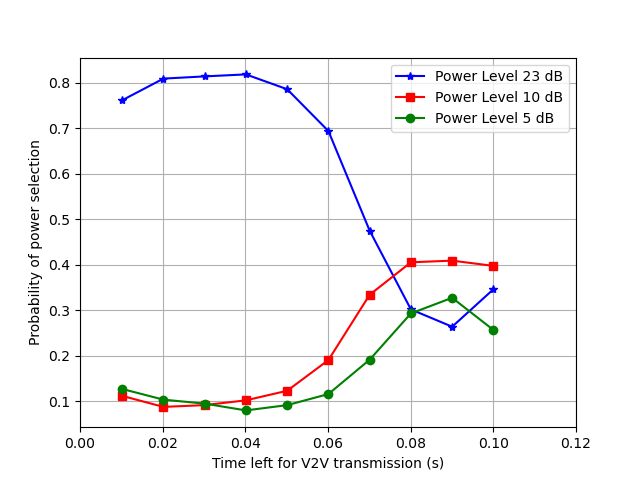}
		\caption{Relationship between remaining time and power selection.}
		\label{fig8}
	\end{figure}

	As shown in Fig. \ref{fig8}, to analyze the impact of GNN on the strategies learned by the agents, we followed the method of \cite{8633948}, collecting the decisions made by the agents under different remaining transmission times and analyzing the statistical probability of each power level being selected. Unlike \cite{8633948}, we imposed a significant penalty on the agents when V2V link transmission failed. Consequently, when the remaining transmission time is only 0.01 seconds, the agents tend to choose higher power levels. As illustrated in the figure, when the transmission time is ample, the agents tend to select lower power levels to reduce interference with V2I links. However, when the transmission time is tight, they tend to choose higher power to ensure the success of V2V link transmission. Additionally, at the beginning of the transmission, agents also tend to select higher power, possibly because the agents do not reselect channels in every time slot, and initially setting a higher power level is a prudent choice. \textcolor{red}{This indicates that the strategies learned by the agents captures certain levels of experience, helping to analyze the usefulness of the strategies.}
	
	\begin{figure}[t]
		\centering
		\includegraphics[width=\columnwidth]{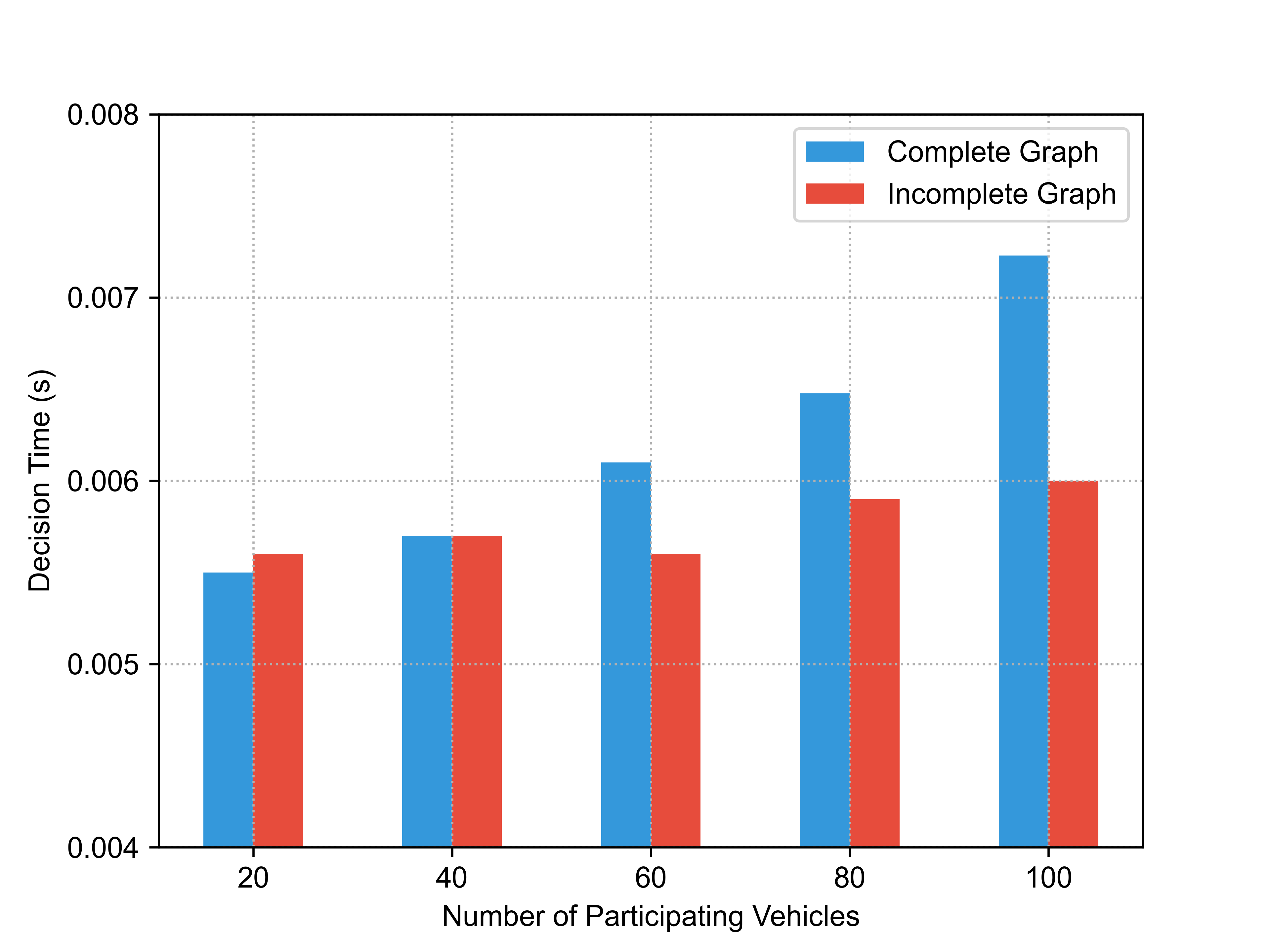}
		\caption{Comparison of single decision time between complete and incomplete graphs.}
		\label{fig9}
	\end{figure}\par
	Fig. \ref{fig9} illustrates the performance differences when constructing the GNN using a complete graph versus an incomplete graph. When using a complete graph, the number of nodes in the environment increases linearly with the number of vehicles, leading to a slight increase in the required training time. \textcolor{red}{It can be observed that the time required for using an incomplete graph remains almost the same as a constant, whereas the time required for a complete graph gradually increases as the number of nodes increases.} To ensure the reliability of the comparison, we omit the data preprocessing steps for the incomplete graph scenario here, as the incomplete graph requires an additional step to obtain neighbors from the graph. We did not employ advanced, efficient APIs optimized for GNN to accelerate this process \cite{Zheng2021Feluca, Zheng2023Path}; instead, we implemented it using the original code. Consequently, processing on the CPU may consume some time, including delays caused by transferring data from CPU to GPU. Our comparison focuses on the computation speed on the GPU. For the incomplete graph curve, since the number of neighbors is small and does not increase with the number of vehicles, the time required for a single decision remains relatively small and is less affected by the number of vehicles in the environment.
	
	\begin{figure}[t]
		\centering
		\includegraphics[width=\columnwidth]{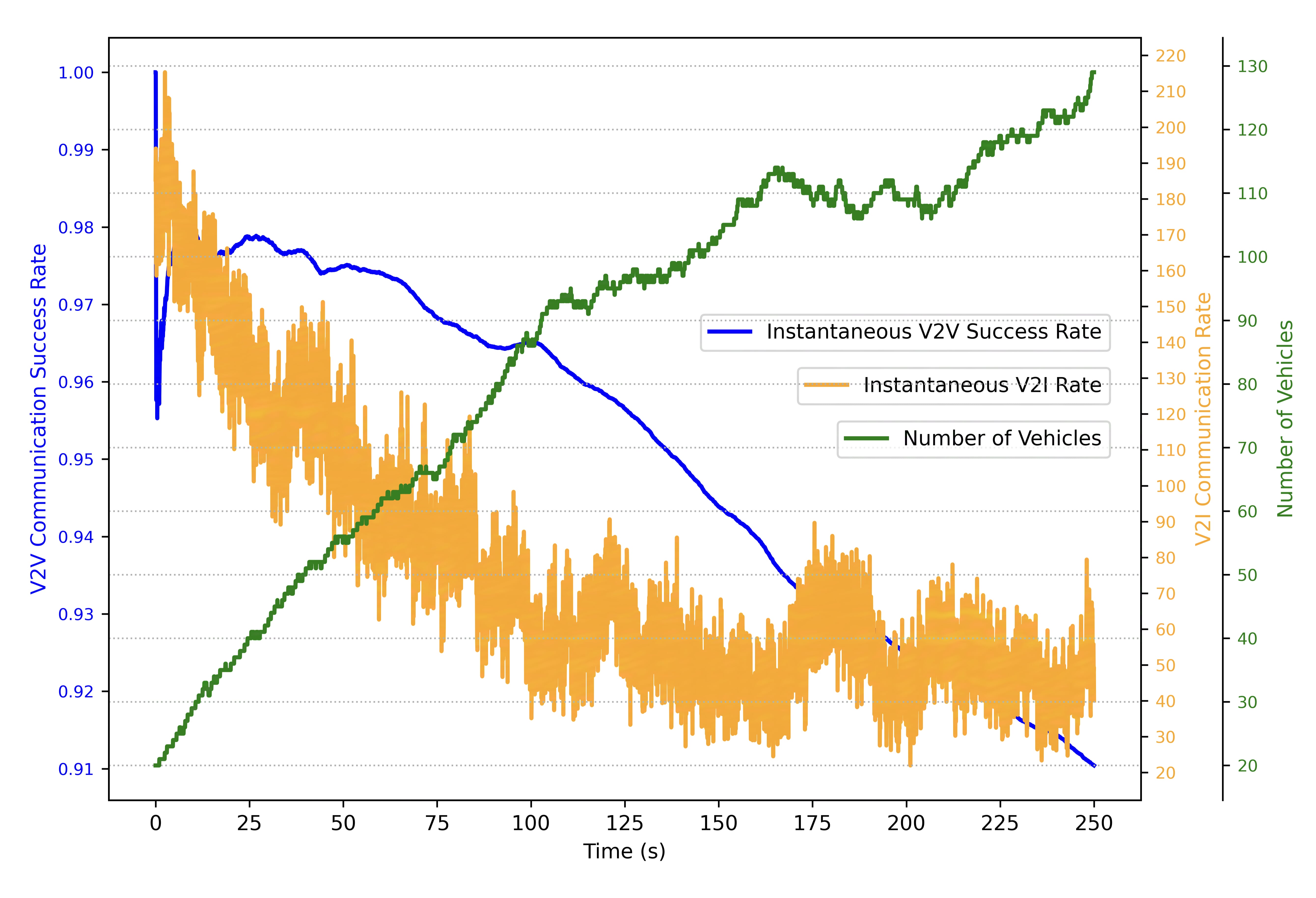}
		\caption{Temporal dynamics of vehicle count, V2I rates, and V2V success rates in dynamic environments.}
		\label{fig10}
	\end{figure}
	
	\begin{figure}[t]
		\centering
		\includegraphics[width=\columnwidth]{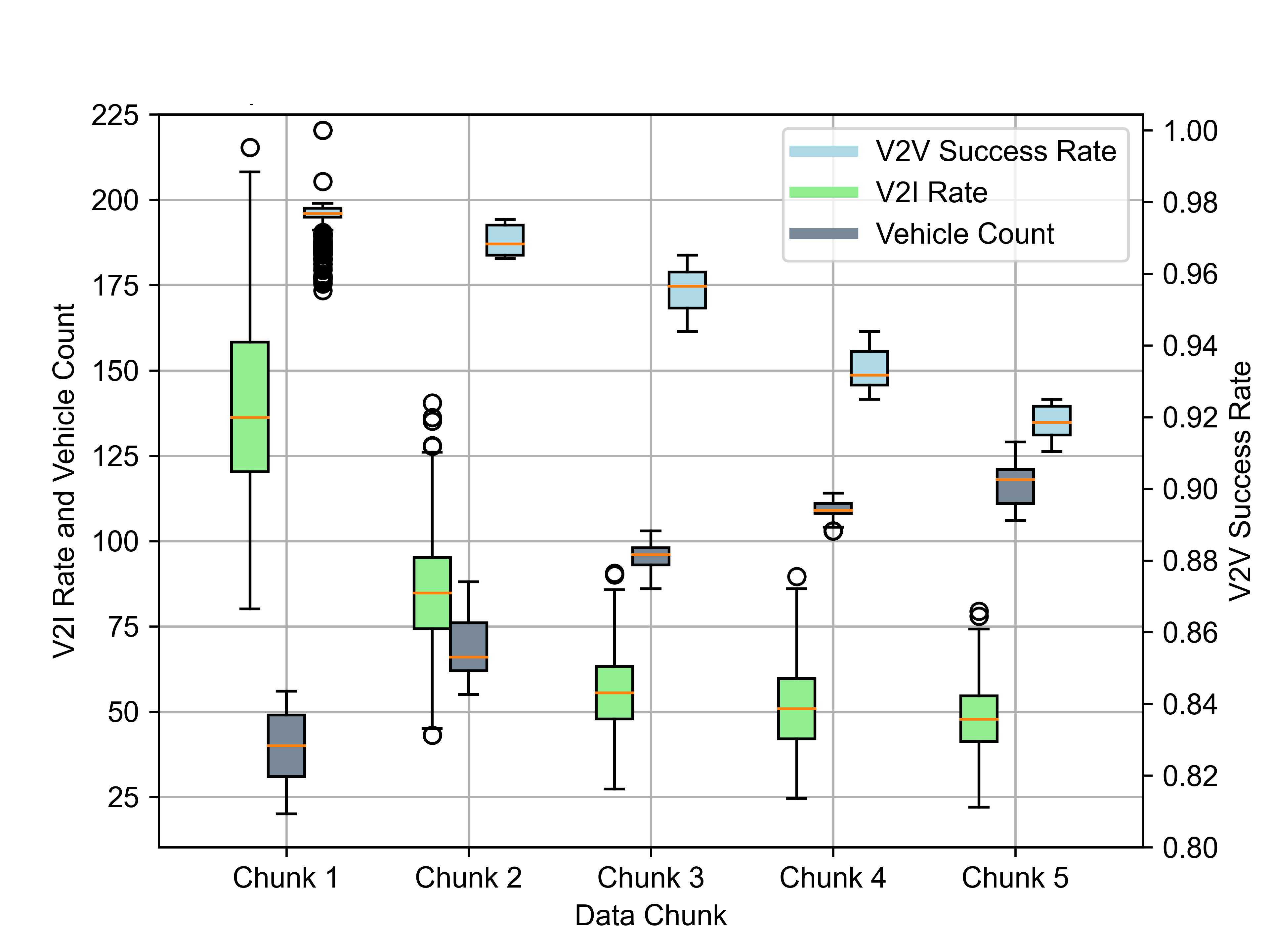}
		\caption{Algorithm performance box plot: vehicle count, V2I rates, and V2V success across five time segments.}
		\label{fig11}
	\end{figure}	
	
	\subsection{Dynamic Environment Testing}
	Finally, to validate the performance of our proposed method in environments with dynamically changing vehicle numbers, we provide the test results under dynamic conditions. \textcolor{red}{On one hand, the dynamic environment requires high robustness from the algorithm as the vehicle distribution frequently changes over time. On the other hand, the rapidly changing channel occupancy in the environment demands higher and robust performance from the algorithm.} To obtain a rich set of test data, we designed a dynamic environment using a dictionary to store the vehicles, deleting the corresponding vehicle entry when a vehicle leaves the environment. Each time vehicle positions are updated, there is a certain probability of adding new vehicles; if no vehicles are added, the probability increases. Once a vehicle is added, the probability is reset to zero. To ensure a thorough exploration of a greater number of vehicles, we adopt a probability mode for vehicle addition, ensuring a steady increase in the number of vehicles within the environment.
	
	In the decision-making process within the dynamic environment, one-tenth of the vehicles in the environment make decisions in each time slot. That is, vehicles are divided into ten batches to make decisions sequentially, completing one batch every 0.005 seconds, consistent with the method we previously proposed.
	
	\begin{itemize}
		\item[1)] The number of vehicles, instantaneous V2I rate, and instantaneous V2V success rate in a dynamic environment: Fig. \ref{fig10} shows the variation over time of the number of vehicles, the instantaneous average success rate of V2V communication, and the sum of the instantaneous V2I communication rates across all channels. The number of vehicles increases continuously over time. At the beginning of the environment, there is a significant oscillation in the V2V communication success rate. This oscillation, which does not occur consistently in every test, is due to the randomness of the environment and quickly stabilizes. As can be seen from the figure, the V2V communication success rate remains very stable and does not fluctuate with changes in vehicle positions, but it gradually decreases as the number of vehicles in the environment increases. For the V2I rate, since the strategy is trained in a static environment, it cannot fully adapt to the changing environment. While ensuring the success rate of V2V communication, it does not effectively minimize interference with the V2I links, resulting in considerable fluctuations in the V2I rate. Overall, the V2I rate shows a downward trend with the increase in the number of vehicles.
		\item[2)] Results Analysis: In Fig. \ref{fig11}, the sampled 5000 data points are divided into five segments, each representing data over a specific time period. The performance of the number of vehicles, V2I communication rate, and V2V communication success rate are analyzed for each time period. During the first time period, the V2V communication success rate exhibits a higher frequency of outliers, which can be attributed to the environmental oscillations observed from the instantaneous values. \textcolor{red}{Although the V2I communication rate occasionally shows outliers, but still be within a controllable range. Overall, as the number of vehicles increases gradully, the V2I rate and the V2V success rate appear to decline gradually. Their values in each time period remain relatively stable with few outliers, demonstrating the robustness of the algorithm.}
	\end{itemize}		
	\section{Conclusions}
	\label{sec7}
	In this paper, we started with the reason that centralized resource allocation methods outperform distributed ones, and attempted to integrate GNN with DRL to enable each agent to acquire more information from local observations. To address the issue that GNNs are typically deployed on a global scale and struggle to adapt to dynamic graph structures, we proposed a novel graph construction method that is perfectly suited for vehicular networks. This method implicitly constructs the graph network without adding extra communication overhead. Additionally, an inductive GraphSAGE model was employed to handle the variability in the number of vehicles. Finally, we combined the proposed GNN model with a DDQN to solve the resource allocation problem, and constructed both a static environment with a fixed number of vehicles and a dynamic environment with a varying number of vehicles to test it. Simulation results demonstrated that our proposed GNN model can effectively improve the policies learned by the agents, achieving better performance that the standalone DQN method. 
	
	Based on theoretical analysis and simulation results, the conclusions can be summarized as follows:
	\begin{itemize}
		\item The graph construction method proposed in this paper ensures distributed network deployment and reduces the decision-making time required.
		\item The GraphSAGE model can capture features that include global structural information from the graph.
		\item The integration of GNN enables intelligent agents to capture more information from local observations, facilitating better decision-making.
	\end{itemize}
	
	\textcolor{red}{Despite the strengths of the proposed method, there are still areas for improvement. The GraphSAGE model's use of an inductive approach to learn a general aggregation function makes it challenging to adapt to the individual characteristics of different nodes, which limits its performance. Future work could focus on enhancing the model architecture to ensure that the GNN can adapt to variations in the number of vehicles while extracting more precise state features. This could involve developing more sophisticated aggregation functions or incorporating additional contextual information into the model. By addressing these challenges, the proposed method can achieve even better performance and further optimize resource allocation in vehicular networks.}
				

				
	\ifCLASSOPTIONcaptionsoff
	\newpage
	\fi

\end{document}